\documentclass{article}

\usepackage[preprint, nonatbib]{neurips_2020}

\usepackage[utf8]{inputenc} % allow utf-8 input
\usepackage[T1]{fontenc}    % use 8-bit T1 fonts
\usepackage{url}            % simple URL typesetting
\usepackage{booktabs}       % professional-quality tables
\usepackage{amsfonts}       % blackboard math symbols
\usepackage{nicefrac}       % compact symbols for 1/2, etc.
\usepackage{microtype}      % microtypography

\usepackage{algorithm}
\usepackage{algorithmicx}
\usepackage{multirow}
\usepackage{amsmath}
\usepackage{stmaryrd}
\usepackage{float}
\usepackage{graphicx}
\usepackage{subfigure}
\usepackage{xcolor}
\usepackage{stmaryrd}
\SetSymbolFont{stmry}{bold}{U}{stmry}{m}{n}

\newtheorem{theorem}{Theorem}%[section]
\newtheorem{lemma}[theorem]{Lemma}

\newcommand{\ex}{{\bf\sf E}}

\newcommand{\boldone}{\mbox{$\boldsymbol{1}$}}
\newcommand{\boldx}{\mbox{$\boldsymbol{x}$}}
\newcommand{\boldy}{\mbox{$\boldsymbol{y}$}}
\newcommand{\boldu}{\mbox{$\boldsymbol{u}$}}
\newcommand{\boldv}{\mbox{$\boldsymbol{v}$}}

\newcommand{\boldeta}{\mbox{$\boldsymbol{\eta}$}}

\newcommand{\prob}{\mbox{\rm Prob}\,}

\newcommand{\var}{{\bf\sf Var}}

\newcommand{\RR}{\mathbb R}

\newcommand{\Rmnum}[1]{\uppercase\expandafter{\romannumeral #1}} % Roman capital letter
\usepackage[raggedright]{titlesec}% can be changed to raggedleft  raggedright
\titleformat{\chapter}{\centering\Huge\bfseries}{Chapter \Rmnum{\thechapter} }{1em}{} % in book format
\usepackage{mathrsfs} %math-script format, then use \mathscr{F}
\usepackage{enumerate} % change the symbols in enumerate
\graphicspath{{figures/}}

\title{ Binary Random Projections with Controllable Sparsity Patterns }

\author{%
  Wenye Li\\
  The Chinese University of Hong Kong, Shenzhen\\
  Shenzhen, Guangdong, China \\
  \texttt{wyli@cuhk.edu.cn} 
  \And
  Shuzhong Zhang\\
  University of Minnesota\\
  Minneapolis, MN 55455, USA \\
  \texttt{zhangs@umn.edu} 
}

\begin{document}

\maketitle

\begin{abstract}
Random projection is often used to project higher-dimensional vectors onto a lower-dimensional space, while approximately preserving their pairwise distances. It has emerged as a powerful tool in various data processing tasks and has attracted considerable research interest. Partly motivated by the recent discoveries in neuroscience, in this paper we study the problem of random projection using binary matrices with controllable sparsity patterns. Specifically, we proposed two sparse binary projection models that work on general data vectors. Compared with the conventional random projection models with dense projection matrices, our proposed models enjoy significant computational advantages due to their sparsity structure, as well as improved accuracies in empirical evaluations.
\end{abstract}

\section{Introduction}
\label{sec:intro}

Random projection is a powerful tool in reducing the dimensionality of a set of points in Euclidean space while approximately preserving their pairwise distances. The method is simple to implement, alongside its strong theoretical guarantees (cf.~\cite{johnson1984extensions,vempala2005random}). Moreover, the method has demonstrated superior practical performances in a wide variety of empirical applications, such as in nearest neighbors search, natural language representation, information retrieval, and image processing;  see~\cite{kanerva2000random,bingham2001random,manning2008introduction,leskovec2020mining}. In practice, one popular choice of the projection matrix is random dense matrix with the entries, say, coming from i.i.d.\ normal distributions. However, it is also possible to use random but {\em sparse}\/ projection matrices, which have been reported to achieve significantly improved computational efficiency with little or no accuracy losses; see~\cite{achlioptas2003database,dasgupta2010sparse,kane2014sparser,dasgupta2017neural}.

In this paper, we aim to investigate random projection using sparse binary matrices with {\it controllable}\/ sparsity pattern, partly motivated by the recent discoveries in neuroscience \cite{lin2014sparse,zheng2018complete}. However, a na\"{i}ve application of binary matrix (with $0$-$1$ entries) as the projection matrix does not guarantee the distance-preserving property. As a remedy, we propose two random projection models that eliminate the statistical bias in the existing approaches. Compared with the classical random projections with random normal matrices, our newly proposed methods are not only simple in producing sparse solutions but also have a smaller variance in the norm of vector being projected; therefore, it has an improved ability to preserve pairwise distances between the points during projection.

The paper is organized as follows. Section \ref{sec:related} introduces the related work. Section \ref{sec:model} presents our proposed model. Section \ref{sec:eval} provides the evaluation results, followed by conclusion in Section \ref{sec:conc}. Besides, some proofs of key results are delegated to the appendix of the manuscript.

%%%%%%%%% Section 2
\section{Related Work}
\label{sec:related}

\subsection{Random projection}
\label{sec:related:rp}

Let $V\in \RR^{d\times n}$ be a matrix whose columns represent $n$ points in the $d$-dimensional real Euclidean space. The random projection method, which multiplies a random matrix 
$W\in \RR^{m\times d}$ ($m\ll d$) with $V$, 
reduces the dimension of the ambient space from $d$ down to $m$. The matrix $W$ typically consists of entries generated from standard normal distribution $N\left(0,1\right)$. A classical result, known as the Johnson--Lindenstrauss theorem, states that random projection can well preserve pairwise distances between the original points in $\RR^{d}$ with high probability; see \cite{johnson1984extensions,allen2014sparse,larsen2017optimality}.

Further computational speedups can be achieved through sparsification of the projection matrix. In \cite{achlioptas2003database,li2006very}, the authors propose sparse random projections by replacing the $N\left(0,1\right)$ entries in $W$ with entries taken from $\left\{+1,0,-1\right\}$ with specifically designed probabilities. In \cite{bourgain2015toward}, a sparse model that treats non-zero entries as the Rademacher variables is proposed. Lower bounds on the sparsity of the projection matrix in order to preserve Euclidean distances for all points have also been established in \cite{ailon2009fast,kane2014sparser,jagadeesan2019understanding}. These methods have significantly improved the projection efficiency with little sacrifice in loss of accuracy; therefore they are used in large-scale data processing applications \cite{bingham2001random}.

\subsection{Sparse binary projection}
\label{sec:related:sbp}

There is a general framework to map $d$-dimensional vectors to $m$-dimensional ones with a sparse binary random projection matrix, which we shall call {\it sparse binary projection}\/ in this paper. Specifically, the projection matrix $W_{B}=\left(\xi_{ij}\right)_{m\times d}$ is constructed with each entry $\xi_{ij}$ being generated as i.i.d.\ Bernoulli random variable:
\[
\xi_{ij} = \left\{
\begin{array}{rl}
1, & \mbox{with probability $p$}, \\
0, & \mbox{with probability $\left(1-p\right)$},
\end{array}
\right.
\]
where $0<p \le 1/2$, $i=1,...,m$ and $j=1,...,d$.

Let two vectors $\boldu,\boldv\in \RR^{d}$ be projected to $\boldu',\boldv'\in \RR^{m}$ with the projection matrix $W_{B}$. It is easy to show (see e.g.\ \cite{dasgupta2017neural}) that
\begin{lemma}
Take any $\boldu,\boldv\in \RR^{d}$ and define $\boldu'=\frac{1}{\sqrt{m p (1-p)}} W_{B} \boldu$ and $\boldv'=\frac{1}{\sqrt{m p (1-p)}} W_{B} \boldv$. Then
\[
\ex \left( \| \boldu'-\boldv'\|^2 \right)
= \| \boldu-\boldv \|^2 + \frac{p}{1-p} [\boldone\cdot (\boldu-\boldv)]^2
\]
where $\boldone$ is the all-ones vector (and thus $\boldone\cdot \boldu$ is the sum of the entries of $\boldu$).
\end{lemma}

The sparse binary projection model works when all input vectors have been centered through divisive normalization \cite{olsen2010divisive,papadopoulou2011normalization}, i.e., $\boldsymbol{1}\cdot \boldu \approx \boldsymbol{1}\cdot \boldv$ for all pairs of vectors. In such cases, the term 
$\frac{p}{1-p} \left[\boldsymbol{1}\cdot \left(\boldu-\boldv\right)\right]^2 \approx 0$ 
and is thus negligible. Therefore the distance between $\boldu'$ and $\boldv'$ can be used to approximate the distance between $\boldu$ and $\boldv$. However, for general input vectors, $\| \boldu'-\boldv'\|^2$ does not necessarily preserve the distance due to the existence of the second error term, which is non-negligible and it brings a bias to the estimation.

In addition to the model defined above, which requires each matrix entry be set independently with the uniform probability, a related model is to let each row of the binary matrix have the same number of ones. Similarly, random projections with such a fixed-ones matrix do not guarantee the distance-preserving property for general data either.

It is worth pointing out that sparse binary projections have a close connection with neural circuits, when forming the tag of an input signal to be used in the mushroom bodies of insect brains for behavioral responses; see \cite{lin2014sparse,stevens2015fly,zheng2018complete}. The connection leads to the development of a novel FLY algorithm \cite{dasgupta2017neural}, which has been reported to yield excellent performances in similarity search tasks.

%%%%%%%% Section 3
\section{Distance-Preserving Sparse Binary Projection Models}
\label{sec:model}

We shall now further study the two sparse binary projection models discussed in Section \ref{sec:related:sbp}, aiming to ensure guaranteed distance-preserving properties on general input data.

\subsection{Bernoulli projection matrix}
\label{sec:model:bernoulli}

For a given $0< p \le 1/2$, let an $m\times d$ binary projection matrix $W_{B}=\left(\xi_{ij}\right)_{m\times d}$ be defined as in Section \ref{sec:related:sbp}. Consider a given $\boldx=\left(x_1,\cdots,x_d\right)^{\top} \in \RR^d$, and denote $\boldeta=\left(\eta_1,\cdots,\eta_m\right)^{\top} \in \RR^{m}$ with
\begin{equation}
\boldeta := \left(W_{B} - p E\right)\boldx = W_{B} \boldx - p E \boldx
\label{equ:matrix_bernoulli}
\end{equation}
where $E\in \RR^{m\times d}$ is the all-ones matrix. It follows that, for $1\le i\le m$,
\[
\eta_{i} := \sum_{k=1}^d \left(\xi_{ik}-p\right) x_k  = \sum_{k=1}^d \xi_{ik} x_k - p (\boldsymbol{1}\cdot \boldx), 
\]
and
\[
\eta_{i}^{2}
= \sum_{k=1}^d (\xi_{ik}-p)^2 x_{k}^{2} + 2 \sum_{1\leq k<\ell \leq d} (\xi_{ik}-p) (\xi_{i\ell}-p) x_{k}x_{\ell}.
\]
Therefore, the expectations of each $\eta_{i}^{2}$ and $\| \boldeta \|^{2}$ are:
\begin{equation}
\ex (\eta_{i}^{2})
= \sum_{k=1}^d p\left(1-p\right) x_k^{2}
= p\left(1-p\right) \|\boldx\|^{2},
\quad
\ex (\| \boldeta \|^{2})
=\sum_{i=1}^{m}\ex(\eta_{i}^{2})
= m p \left(1-p\right) \|\boldx\|^{2}.
\label{equ:e_eta_2}
\end{equation}

It can be computed that
\begin{equation}
\ex (\eta_{i}^{4})
= p (1-p)(1-6p+6p^2)\|\boldx\|_{4}^{4} + 3 p^2\left(1-p\right)^2\|\boldx\|^{4},
\label{equ:e_eta_4}
\end{equation}
where $\|\boldx\|_{4}^{4}=\sum_{k=1}^{d}x_{k}^{4}$ and $\|\boldx\|^{4}=\left(\sum_{k=1}^{d}x_{k}^{2}\right)^{2}$.

\subsection{Concentration bounds}
\label{sec:model:bernoulli_bound}

Next, we estimate how tightly the estimate $\|\boldeta\|^2$ is concentrated around its expectation. Denote $\zeta_i := \eta_i^2 $, for $i=1,2,...,m$. Assume that $0<p\le 1/2$. Because $\left|\eta_i\right| \le  \left(1-p\right)\sqrt{d} \|\boldx\| $
for $1\le i \le m$, due to Cauchy-Schwarz inequality, we have
\begin{equation}
\zeta_i \le (1-p)^2 d \|\boldx\|^{2} \mbox{, for $i=1,2,...,m$}.
\label{equ:zeta}
\end{equation}

By \eqref{equ:e_eta_4}, we conclude that
\begin{equation}
\ex (\zeta_{i}^{2}) < 2\left(1-p\right)^{3} \|\boldx\|^{4} \mbox{ for $i=1,...m$}.
\label{equ:e_zeta_2}
\end{equation}

Note that the random variables $\zeta_i$'s are independent to each other, which makes it possible to derive a concentration bound using a slightly modified version of Bennett's inequality \cite{bennett1962probability}:

\begin{lemma} \label{bennett} (Bennett's inequality)
Let $Z_{1},\cdots ,Z_{m}$ be independent nonnegative random variables with finite variances and that $Z_{i} \leq b$ for some $b>0$ almost surely for all $i=1,2,...,m$. Let
\(
S=\sum_{i=1}^{m}\left[ Z_{i}-\ex (Z_{i})\right]
\)
and $w \geq \sum_{i=1}^{m} \ex (Z_{i}^{2})$. Then we have, for any $t>0$,
\[
\prob \left\{ S\geq t\right\} \leq \exp \left( -\frac{w}{b^{2}}h\left( \frac{bt}{w}\right) \right)
\]
where $h\left( a\right) =\left( 1+a\right) \log \left( 1+a\right) -a$ for $a>0$.
\end{lemma}

In our case, we may let $b=\left( 1-p\right)^{2} d \|\boldx\|^{2}$ (cf.~\eqref{equ:zeta}), $w = 2m \left(1-p\right)^{3} \|\boldx\|^{4}$ (cf.~\eqref{equ:e_zeta_2}), and let
$t=\epsilon m p \left(1-p\right) \|\boldx\| ^{2}$ (cf.~\eqref{equ:e_eta_2}).
Notice that when $0<a\leq 4$, $\frac{a^2}{4} < h(a)$. So when $0<\epsilon \leq\frac{8}{d p}$, we have $\frac{b t}{w}\leq 4$, and so $\frac{1}{4}\left( \frac{b t}{w}\right) ^{2}<h\left( \frac{b t}{w}\right)$.
Applying Lemma~\ref{bennett}, we obtain
\[
\prob \left\{ \sum_{i=1}^{m}\left[ \zeta_{i}-\ex(\zeta_{i})\right] 
\geq \epsilon \sum_{i=1}^m \ex (\zeta_{i}) \right\}
<\exp \left( -\frac{t^2}{4w}\right)
=\exp \left( -\frac{\epsilon ^{2}m p^{2}}{8\left(1-p\right)}\right) .
\]

The result from Bennett's inequality provides the upper tail of the bound. The other side of the bound can be obtained from a well-known result on sub-Gaussian lower tails for nonnegative random variables (cf.~page 47 of \cite{boucheron2013concentration}):

\begin{lemma}
Let $Z_{1},\cdots,Z_{m}$ be independent nonnegative variables. Then, for any $t>0$, it holds that
\[
\prob \left\{S\le -t\right\}\le \exp\left(-\frac{t^2}{2w}\right)
\]
where $S=\sum_{i=1}^{m}\left[Z_{i}-\ex (Z_{i})\right]$ and $w\ge\sum_{i=1}^{m} \ex (Z_{i}^{2})$.
\end{lemma}

Applying the above lemma to our setting, we obtain the following bound:
\[
\prob \left\{ \sum_{i=1}^{m}\left[ \zeta_{i}-\ex(\zeta_{i})\right] \leq
-\epsilon \sum_{i=1}^m \ex (\zeta_{i}) \right\}
\leq \exp \left( -\frac{\epsilon ^{2}m p^{2}}{4\left(1-p\right)}\right) .
\]

We scale the projection matrix and let
\[
\boldeta_{B} := \frac{1}{\sqrt{m p\left(1-p\right)}} \left(W_{B} - p E \right)\boldx.
\]
Furthermore, by scaling relative to the error we finally arrive at a probability bound on both sides:
\begin{equation}
\prob \left\{
\left(1-\epsilon \right)\| \boldx\|^{2}
\le \| \boldeta_{B} \|^{2}
\le \left(1+\epsilon\right)\| \boldx\|^{2} \right\}
> 1-2\times \exp\left( -\frac{\epsilon ^{2}m p^{2}}{8\left(1-p\right)}\right),
\label{equ:bound_bernoulli}
\end{equation}
for $0<\epsilon \leq \frac{8}{d p}$.
Now, let us summarize our findings above as a Johnson--Lindenstrauss style theorem.
\begin{theorem}
Take any $n$ distinct vectors $\boldv_{1},\cdots,\boldv_{n}\in \RR^{d}$. Choose $0<p\leq\frac{1}{2}$ and $0<\epsilon\leq\frac{8}{d p}$. Then, for any
\[
m \ge \frac{16\left(1-p\right) \log n}{\epsilon^{2} p^{2}},
\]
there exists a binary projection matrix $W_{B}$, so that for $\boldv_{i}':=\frac{1}{\sqrt{m p (1-p)}}\left(W_{B} - p E\right)\boldv_{i}$ for $i=1,...,n$, it holds that
\[
\left(1-\epsilon\right)\| \boldv_i-\boldv_j \|^2 < \| \boldv_i'-\boldv_j' \|^2 < \left(1+\epsilon\right)\| \boldv_i-\boldv_j\|^2
\]
for any $1\le i < j\le n$.
\end{theorem}

%%%%%%%%%%% begin proof %%%%%%%%%%%%%%%%
\noindent {\sf Proof.}
For any given pair $i < j$, by setting $\boldx=\boldv_i-\boldv_j$ it follows from \eqref{equ:bound_bernoulli} that
\[
 \prob \left\{  | \| \boldv_i'-\boldv_j'\|^2 - \| \boldv_i-\boldv_j\|^2 | \ge \epsilon \| \boldv_i-\boldv_j\|^2  \right\}
 < 2\times \exp\left( -\frac{\epsilon ^{2}m p^{2}}{8\left(1-p\right)}\right).
\]
By the union bound, if $ m\ge \frac{16\left(1-p\right) \log n}{\epsilon^{2} p^{2}}$ we have
\begin{eqnarray*}
 & &  \prob \left\{  | \| \boldv_i'-\boldv_j'\|^2 - \| \boldv_i-\boldv_j\|^2 | < \epsilon \| \boldv_i-\boldv_j\|^2, \mbox{ for all $1\le i < j \le n$}  \right\} \\
&>& 1 - \frac{n(n-1)}{2} \times 2 \times \exp\left( -\frac{\epsilon ^{2}m p^{2}}{8\left(1-p\right)}\right) \\
&\ge& 1 - n(n-1) \times \exp ( - 2 \log n) \\
&=& \frac{1}{n}.
\end{eqnarray*}
Therefore the required binary projection matrix $W_{B}$ is assured to exist.
\hfill $\Box$
%%%%%%%%%%%% end of proof %%%%%%%%%%%%%%%

\subsection{Fixed-sparsity projection matrices}
\label{sec:model:fixed}

Next we investigate the case of choosing a fixed number of ones in each row of the binary projection matrix. Specifically, we choose $c$ (integer) out of $d$ ($c\leq \frac{d}{2}$) elements in each row to be ones. We model each row of the matrix as complying with a $\binom{d}{c}$-dimensional categorical distribution. Consider an experiment of randomly selecting $c$ elements from a $d$-dimensional vector $x$. There are $\binom{d}{c}$ different combinations. Each combination is denoted by an index set 
$J_{k}\subset \left\{ 1,\cdots,d\right\} $ with
$|J_{k}|=c$ and $1\le k\le \binom{d}{c} $. 
Repeat experimenting $m$ trials and denote the results by 
$\xi_{i}$ ($i=1,\cdots,m$), 
with $\xi_{i}=k$ indicating the $k$-th combination is selected in the $i$-th trial. Assume all combinations have the uniform probability of being selected, i.e.,
\[
\prob \left(\xi_{i}=k\right) = \frac{1}{\binom{d}{c}}, \mbox{ for all $1\le i\le m$, and $1\le k\le \binom{d}{c}$}.
\]

Let the random binary matrix be given as $W_{F}=\left(\xi_{ij}\right)_{m\times d}$ with:
\[
\xi_{ij} = \left\{
\begin{array}{rl}
1, & \mbox{if $j\in J_{\xi_{i}}$}, \\
0, & \mbox{otherwise}.
\end{array}
\right.
\]

Consider a given input vector $\boldx\in \RR^d$. Denote the output vector $\boldeta \in \RR^{m}$ with
\[
\boldeta := \left(W_{F} - c q E\right) \boldx = W_{F} \boldx - c q E \boldx
\]
where $q=\frac{1}{d}\left(1+\sqrt{\frac{d-c}{c(d-1)}}\right)$. We have $\ex (\eta_{i}^{2})=\frac{c(d-c)}{d(d-1)}\|\boldx\|^2$ and

\begin{lemma} \label{constant-sparsity-lemma}
For $5\leq c\leq \frac{d}{2}$ and $0<\epsilon \leq\frac{20}{c}$, it follows that
\[
\prob \left\{ \sum_{i=1}^{m}\left\vert \eta _{i}^{2}-\ex (\eta _{i}^{2})\right\vert \geq
\epsilon \sum_{i=1}^{m} \ex (\eta _{i}^{2}) \right\} 
< 2\times \exp \left( -\frac{\epsilon ^{2}mp\left( 1-p\right) ^{2}}{20}\right)
\]
where $p=\frac{c}{d}$.
\end{lemma}

The proof of Lemma~\ref{constant-sparsity-lemma} can be found in the appendix. By the lemma, letting 
$\boldeta_{F} := \sqrt{\frac{d(d-1)}{m c(d-c)}}\left(W_{F} - c q E \right)\boldx$,
for any $0<\epsilon \leq \frac{20}{c}$, we have
\begin{equation}
\prob \left\{
\left(1-\epsilon \right)\|\boldx\|^{2}
\le \| \boldeta_{F} \|^{2}
\le \left(1+\epsilon\right)\|\boldx\|^{2} \right\}
> 1-2\times\exp\left( -\frac{\epsilon ^{2}m p (1-p)^2}{20}\right).
\label{equ:bound_fix}
\end{equation}

Similar to the analysis in Section \ref{sec:model:bernoulli_bound}, based on \eqref{equ:bound_fix} we have the following Johnson--Lindenstrauss type theorem for the class of fixed row-sparsity projection matrices (i.e.\ exactly $c$ out of $d$ components are ones while all others are zeros):
\begin{theorem} \label{constant-sparsity-JL}
Suppose there are $n$ distinct vectors 
$\boldv_{1},\cdots,\boldv_{n}\in \RR^{d}$ and $5\leq c\leq \frac{d}{2}$. Let $p=\frac{c}{d}$ and $q=\frac{1}{d}\left(1+\sqrt{\frac{d-c}{c(d-1)}}\right)$. Suppose $0<\epsilon \leq \frac{20}{c}$ and
\[
m \ge \frac{40 \log n}{\epsilon^{2} p(1-p)^{2}}.
\]
Then, there exists a binary projection matrix $W_{F} \in \RR^{m\times d}$, where each row of $W_F$  contains exactly $c$ ones while others are zeros,
and by letting $\boldv_i':= \sqrt{\frac{d(d-1)}{m c(d-c)}} \left(W_{F} - c q E\right)\boldv_i$ for $i=1,...,n$, we have
\[
\left(1-\epsilon\right)\| \boldv_i-\boldv_j \|^2 < \| \boldv_i'-\boldv_j' \|^2 < \left(1+\epsilon\right)\| \boldv_i-\boldv_j\|^2
\]
for all $1\le i < j\le n$.
\end{theorem}

\subsection{Comparison of variances}
\label{sec:model:moments}

Consider the random projection that multiplies a vector $\boldx\in \RR^{d}$ with a random matrix $W_{G}\in \RR^{m\times d}$ consisting of i.i.d. $N\left(0,1\right)$ entries. Define $\boldeta_{G}:= \frac{1}{\sqrt{m}} W_{G} \boldx$. Here the subscript ``$G$'' indicates the usage of a random Gaussian matrix.

It can be shown, e.g. from \cite{vempala2005random}, about the mean and the variance of $\|\boldeta_{G}\|^{2}$:
\begin{equation}
\ex \left( \| \boldeta_{G} \|^{2} \right) = \|\boldx\|^{2},
\quad \var \left( \| \boldeta_{G} \|^{2} \right) =\frac{2}{m}\|\boldx\|^{4}.
\label{equ:mv_gaussian}
\end{equation}

For $\boldeta_{B} \in \RR^{m}$ defined in Section \ref{sec:model:bernoulli_bound}, we have $\ex \left( \| \boldeta_{B} \|^2 \right) = \|\boldx\|^{2}$. Note that, for any $\boldeta \in \RR^{m}$ that consists of independent variables $\eta_{i}$ ($1\leq i\leq m$), we have $\var \left( \| \boldeta \|^{2} \right) = \sum_{i=1}^{m} \var \left( \eta_{i}^{2} \right) $ and $\var \left( \eta_{i}^{2} \right) = \ex \left( \eta_{i}^{4} \right) - \left[ \ex \left( \eta_{i}^{2} \right) \right]^{2}$. And it is easy to show (by \eqref{equ:e_eta_2} and \eqref{equ:e_eta_4}) that:
\begin{equation}
\var \left( \| \boldeta_{B} \|^2 \right)
= \frac{1}{m}\left[\frac{1}{p (1-p)} - 6\right] \|\boldx\|_{4}^{4}
+ \frac{2}{m}\|\boldx\|^{4} . 
\label{equ:var_bernoulli1}
\end{equation}

The value of $\var(\|\boldeta_{B}\|^2)$ monotonically decreases in $p$ for $0<p\le \frac{1}{2}$. When $\frac{3-\sqrt{3}}{6} < p \le \frac{1}{2}$, then $\frac{1}{m}\left[\frac{1}{p (1-p)} - 6\right] \|\boldx\|_{4}^{4} < 0$, and so in that case,
\begin{equation}
\var \left( \| \boldeta_{B} \|^2 \right) < \frac{2}{m}\|\boldx\|^{4}.
\label{equ:var_bernoulli2}
\end{equation}
Comparing \eqref{equ:var_bernoulli2} with \eqref{equ:mv_gaussian}, we can see that random projection with the Bernoulli matrix and $\frac{3-\sqrt{3}}{6} < p \le \frac{1}{2}$ provides a less volatile estimate than that with the Gaussian projection matrix.

It is also possible to give a closed-form result for $\boldeta_{F} \in \RR^{m}$ defined in Section \ref{sec:model:fixed}. We have $\ex (\|\boldeta_{F}\|^2) = \|\boldx\|^2$ and
\begin{equation}
\var \left( \| \boldeta_{F} \|^2 \right)
=\frac{d^2 (d-1)^2}{m \binom{d}{c} c^2 (d-c)^2} 
\left(\alpha\|\boldy\|^4 + \beta\|\boldy\|_4^4+\gamma s_y^4+\theta \|\boldy\|_3^3 s_y+\lambda\|\boldy\|^2 s_y^2\right)
-\frac{\|\boldx\|^4}{m}
\end{equation}
where $\boldy=\boldx - q E \boldx$, $s_y=\boldsymbol{1}\cdot \boldy$, $\alpha=3\binom{d-2}{c-2}-6\binom{d-3}{c-3}+3\binom{d-4}{c-4}$, $\beta=\binom{d-1}{c-1}-7\binom{d-2}{c-2}+12\binom{d-3}{c-3}-6\binom{d-4}{c-4}$, $\gamma=\binom{d-4}{c-4}$, $\theta=4\binom{d-2}{c-2}-12\binom{d-3}{c-3}+8\binom{d-4}{c-4}$, and $\lambda=6\binom{d-3}{c-3}-6\binom{d-4}{c-4}$.

In Section \ref{sec:eval:vars}, we shall provide numerical computations which show that $\var(\|\boldeta_{F}\|^2)$ is typically smaller than $\var(\|\boldeta_{B}\|^2)$ on real datasets.

%%%%%%%%%%%%%%% Evaluation
\section{Evaluation}
\label{sec:eval}

\subsection{General setting}
\label{sec:eval:setting}

To evaluate the effectiveness of the proposed methods, in this section we shall carry out a series of empirical evaluations. To this end, ten thousand samples are randomly chosen from each of the following datasets, to be used in the subsequent experimental evaluations: 
\begin{itemize}
  \item GLOVE \cite{pennington2014glove}: pre-trained dense word vectors with the GloVe algorithm in $300$ dimensions.
  \item IMAGENET \cite{russakovsky2015imagenet}: images in $1,000$-dimensional visual words quantized from SIFT features, with around $30\%$ features being non-zero.
  \item RCV1 \cite{lewis2004rcv1}: text newswire stories in bag-of-words. For computational efficiency, we keep a vocabulary of $2,000$ most frequently used words, with around $2.9\%$ features being non-zero.
\end{itemize}

We compare our proposed approaches, denoted by BERNOULLI and FIX-SPARSITY respectively, with several popular random projection approaches, including:
\begin{itemize}
  \item GAUSSIAN: an approach that uses a random normal matrix as the projection matrix.
  \item BOURGAIN \cite{bourgain2015toward}: an approach that uses independent matrix rows, each with exactly $c$ non-zero entries of independent Rademacher variables.
  \item PING \cite{li2006very}: an approach that uses entries from $\left\{+1,0,-1\right\}$, with respective probabilities of $\left\{\frac{1}{2\sqrt{d}},1-\frac{1}{\sqrt{d}},\frac{1}{2\sqrt{d}}\right\}$.
  \item ACHLIOPTAS \cite{achlioptas2003database} \footnote{Achlioptas \cite{achlioptas2003database} also suggests all matrix entries of $\left\{+1,-1\right\}$ with probabilities of $0.5$ each, which is equivalent to BERNOULLI with $p=0.5$.}: an approach that uses entries from $\left\{+1,0,-1\right\}$, with respective probabilities of $\left\{\frac{1}{6},\frac{2}{3},\frac{1}{6}\right\}$.
\end{itemize}

The experiments are performed on a server with four CPUs under the MATLAB platform.

\begin{figure*}[!t]
\centering
 \subfigure[GLOVE]{
   \includegraphics[width=1.6in,height=0.9in]{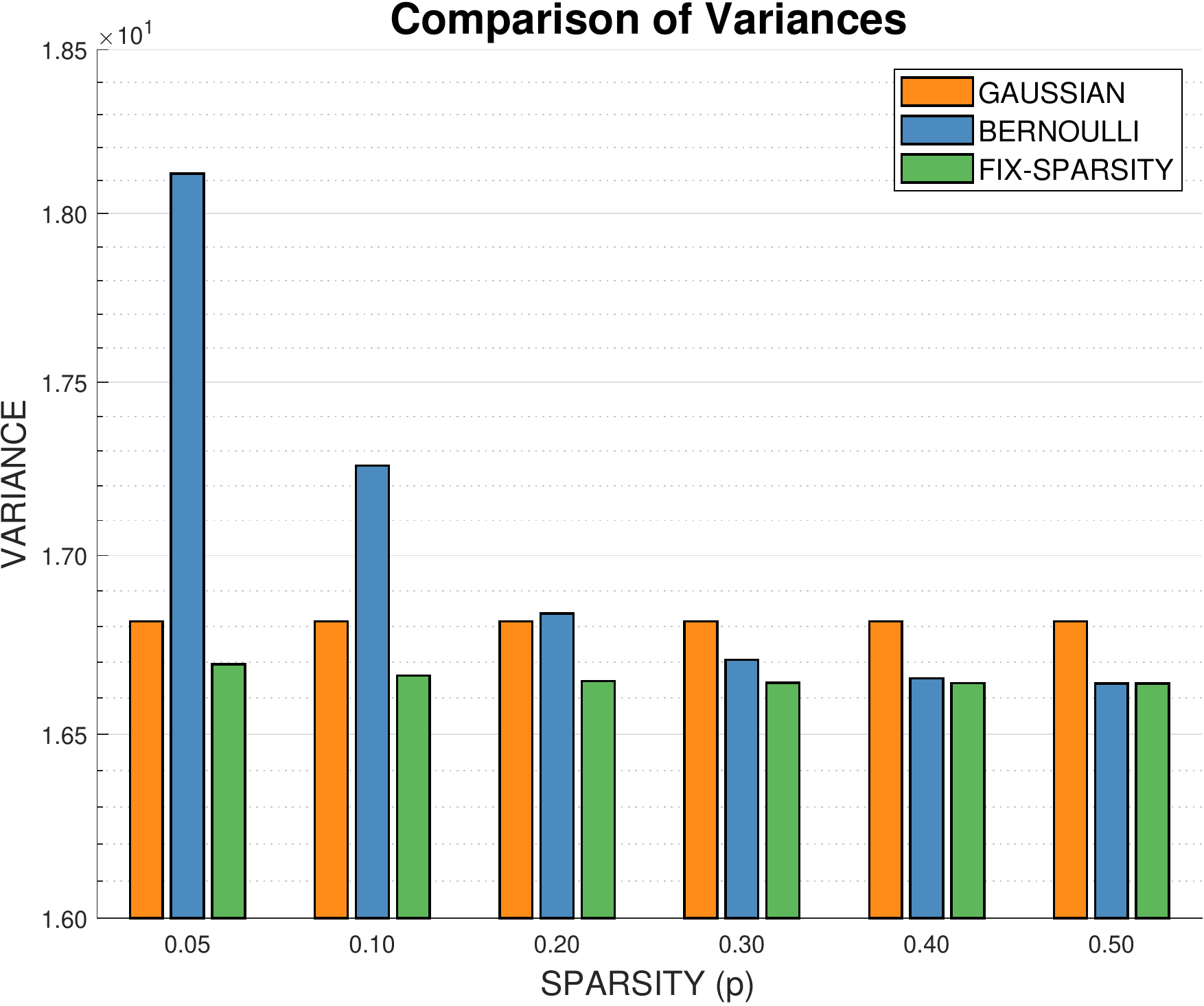}
   \label{fig1:glove}
 }
 \subfigure[IMAGENET]{
   \includegraphics[width=1.6in,height=0.9in]{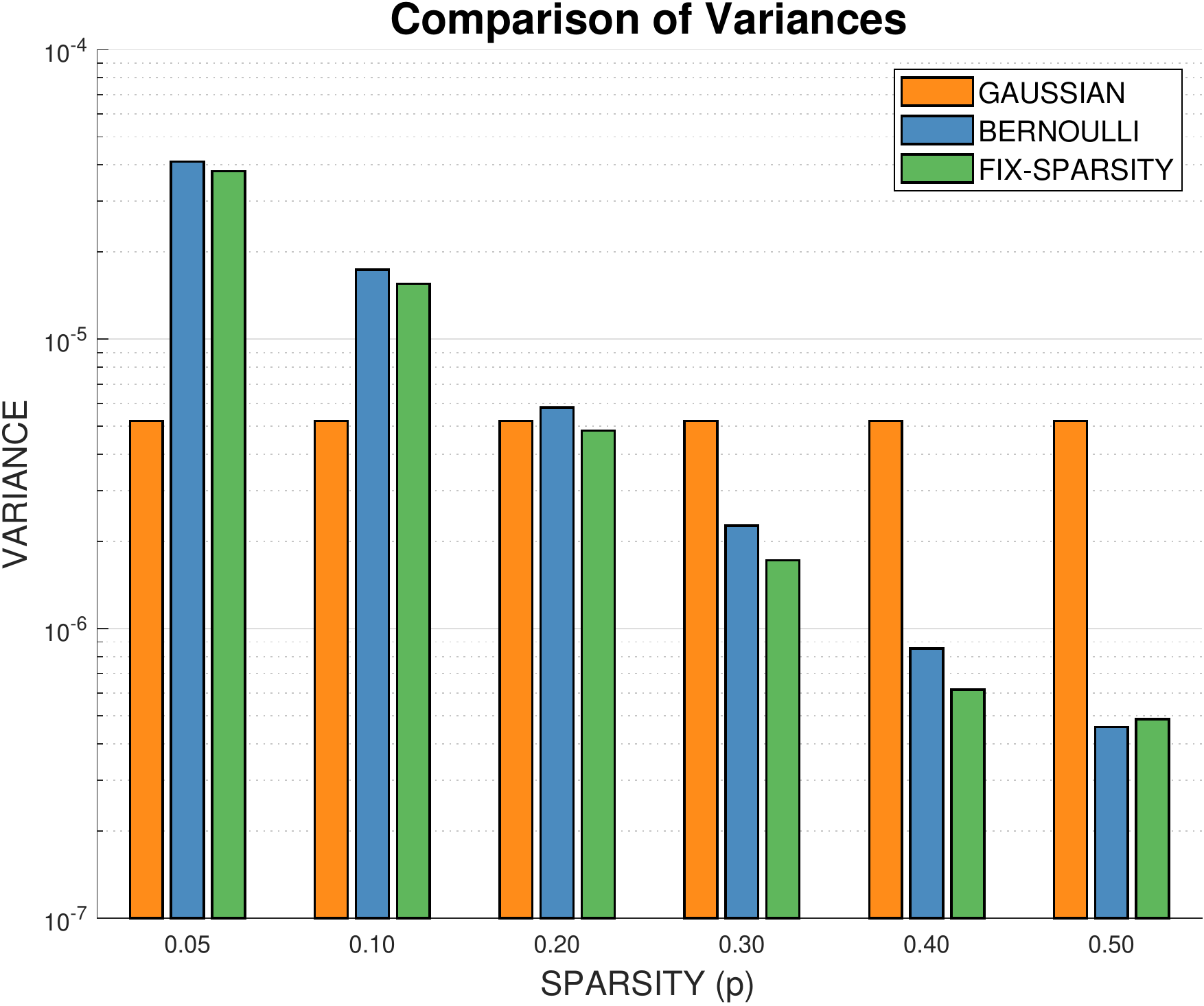}
   \label{fig1:imagenet}
 }
 \subfigure[RCV1]{
   \includegraphics[width=1.6in,height=0.9in]{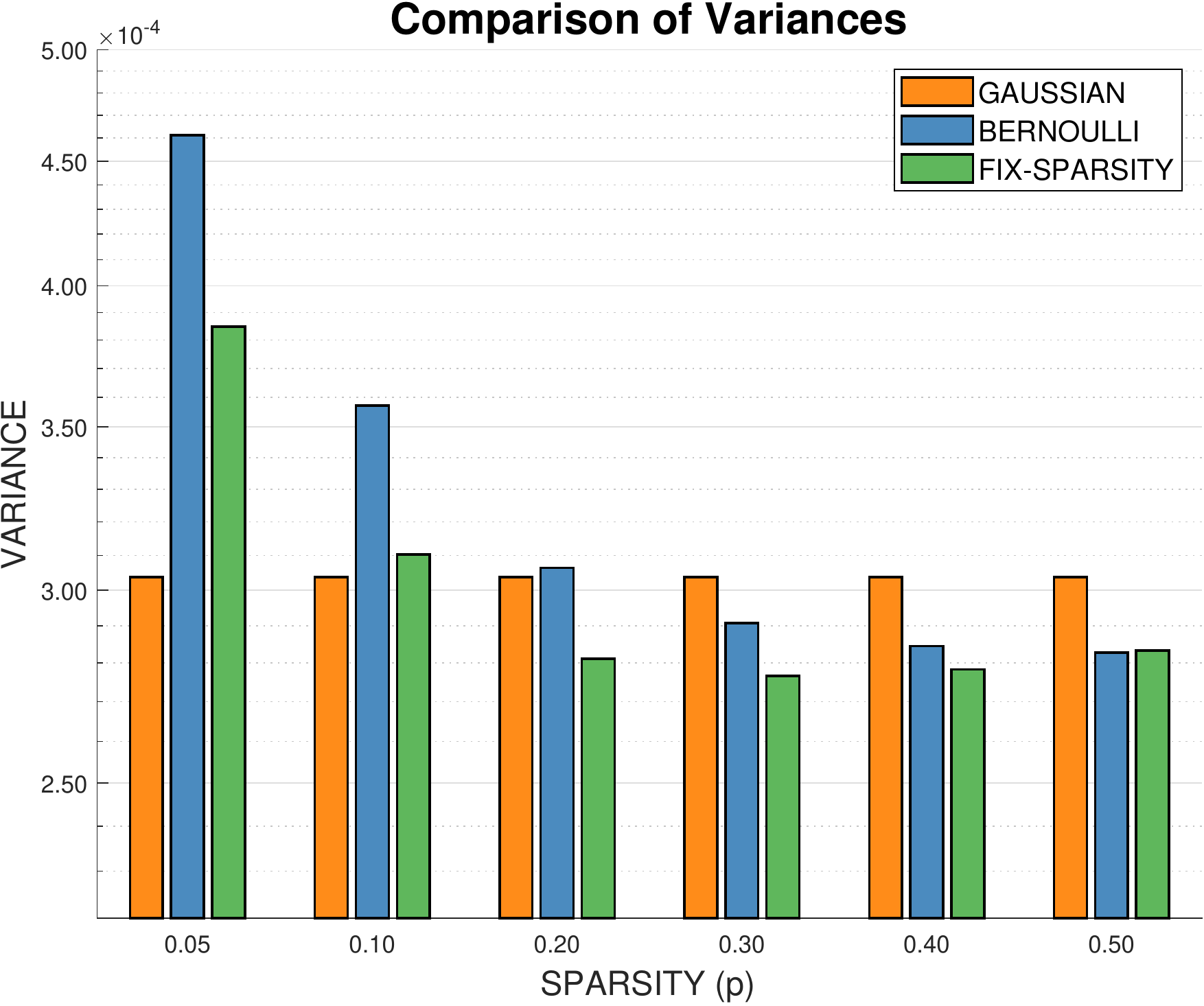}
   \label{fig1:news20}
 }

\caption{\textbf{Comparison of Variances}. X-axis: values of $p$; Y-axis: variances in log-scale. Note: the variances from the Gaussian matrices are not dependent on $p$.}
\label{fig:variances}
\end{figure*}

\begin{figure*}[!t]
\centering
 \subfigure[GLOVE($m=0.1d$)]{
   \includegraphics[width=1.6in,height=0.9in]{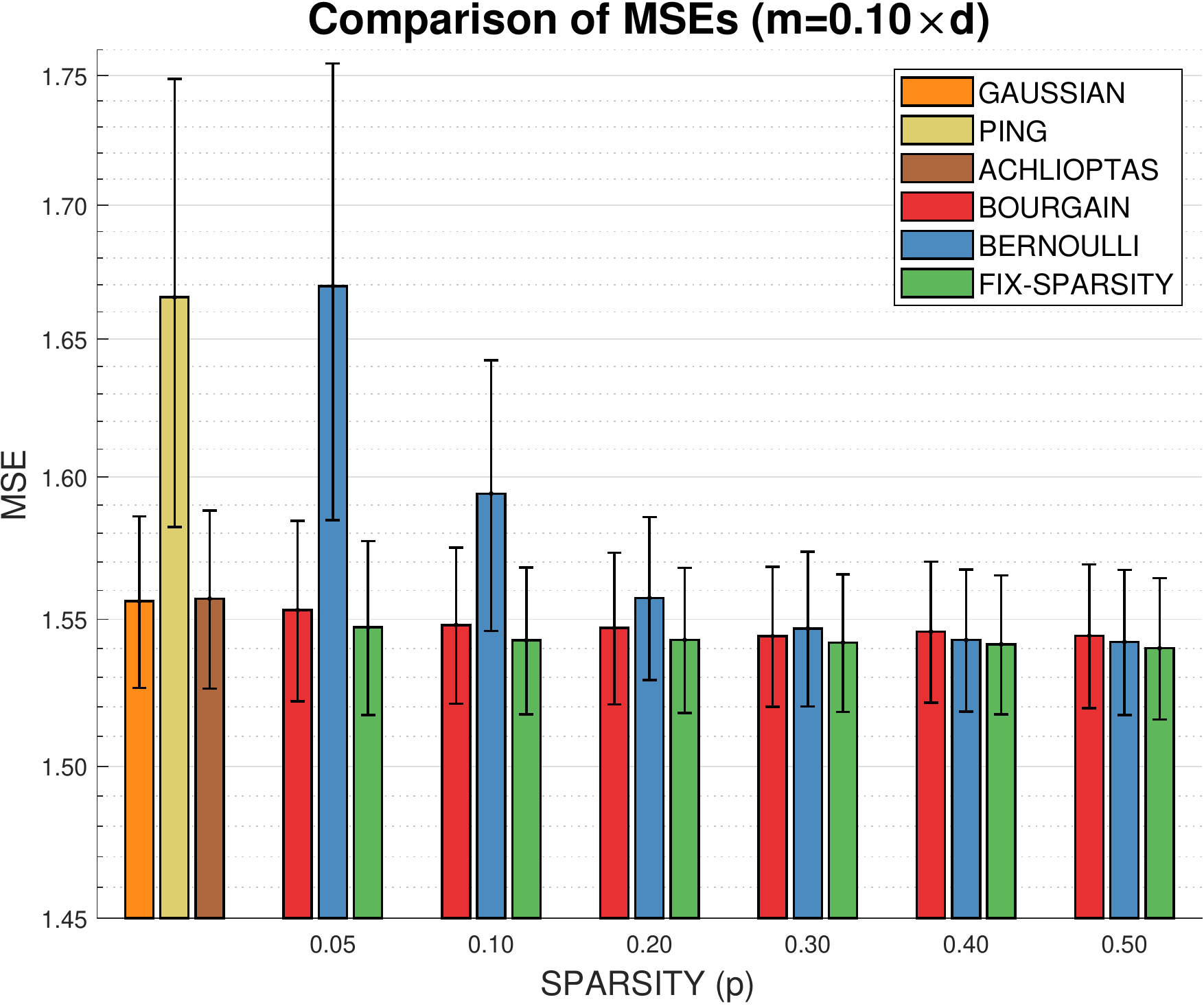}
   \label{fig2:glove:m10}
 }
 \subfigure[IMAGENET($m=0.1d$)]{
   \includegraphics[width=1.6in,height=0.9in]{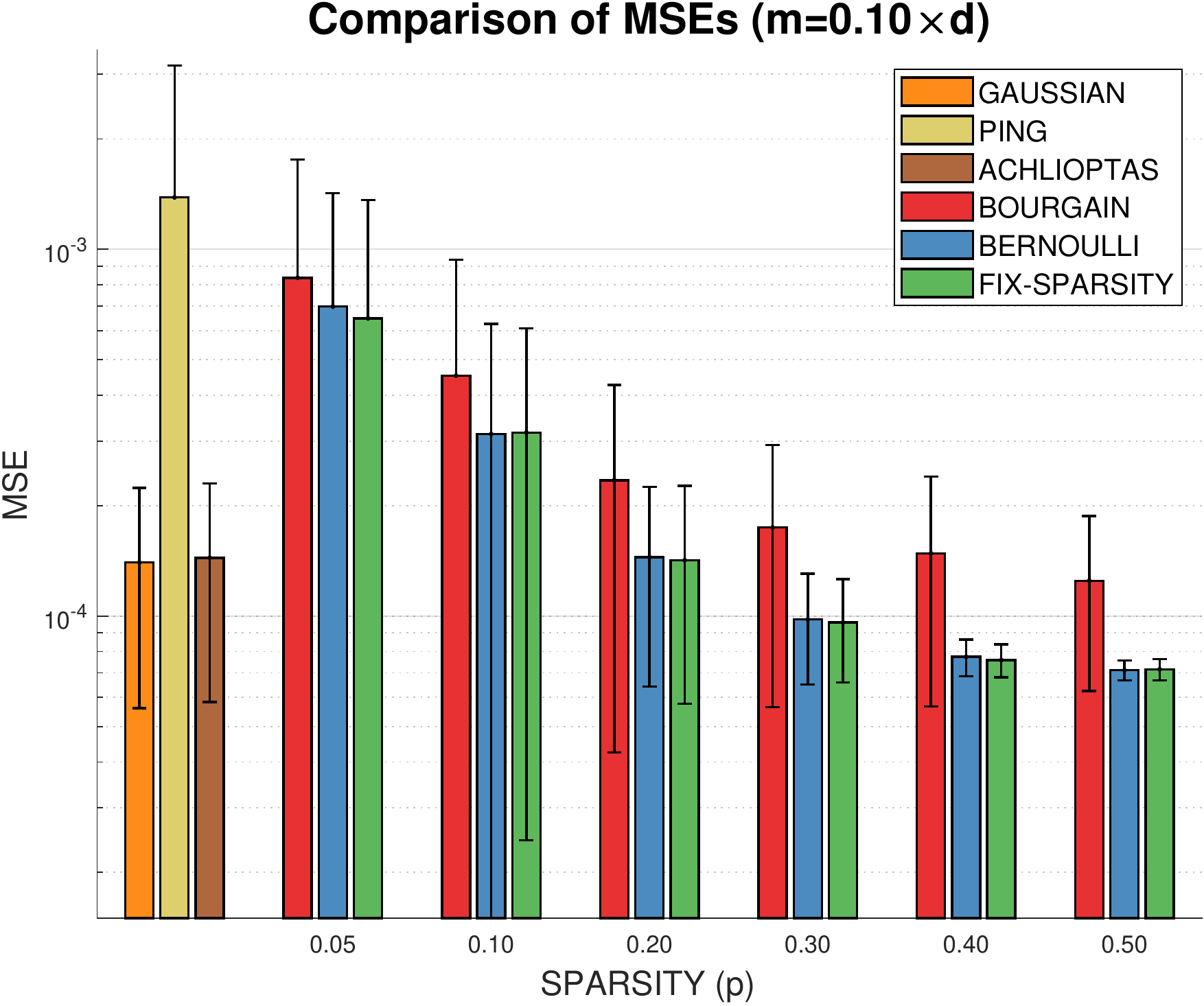}
   \label{fig2:imagenet:m10}
 }
 \subfigure[RCV1($m=0.1d$)]{
   \includegraphics[width=1.6in,height=0.9in]{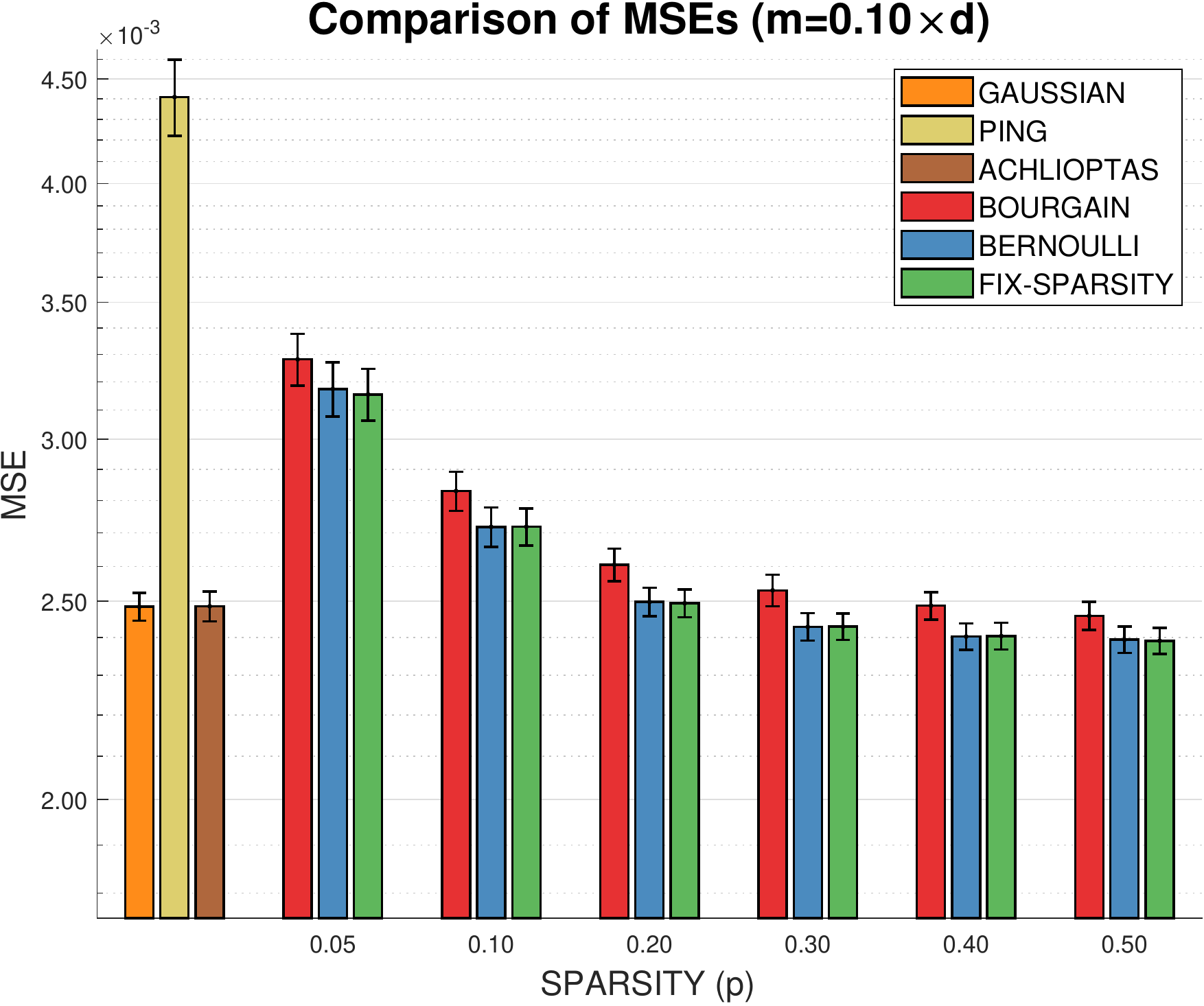}
   \label{fig2:rcv1:m10}
 } \\
 \subfigure[GLOVE($m=0.5d$)]{
   \includegraphics[width=1.6in,height=0.9in]{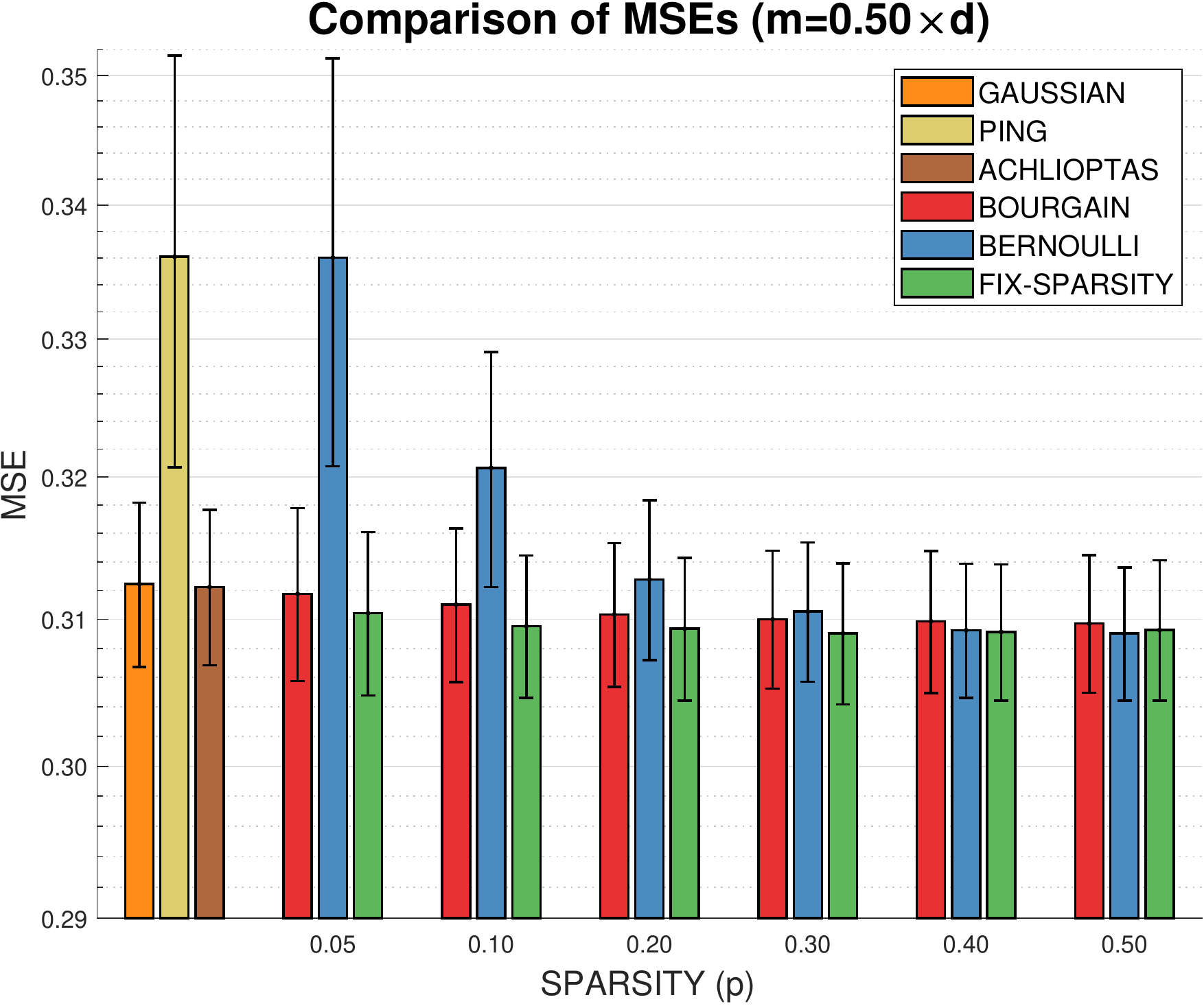}
   \label{fig2:glove:m50}
 }
 \subfigure[IMAGENET($m=0.5d$)]{
   \includegraphics[width=1.6in,height=0.9in]{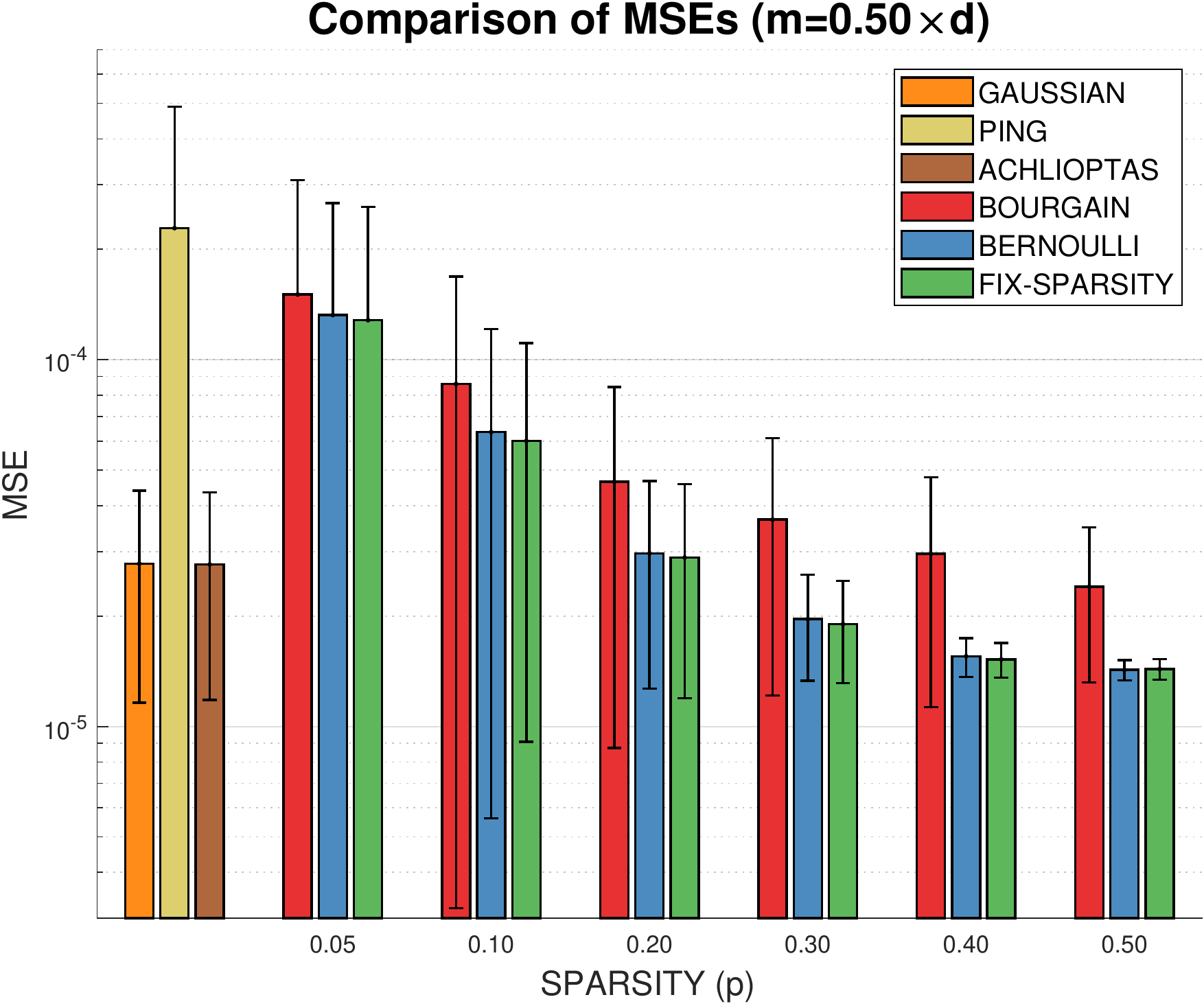}
   \label{fig2:imagenet:m50}
 }
 \subfigure[RCV1($m=0.5d$)]{
   \includegraphics[width=1.6in,height=0.9in]{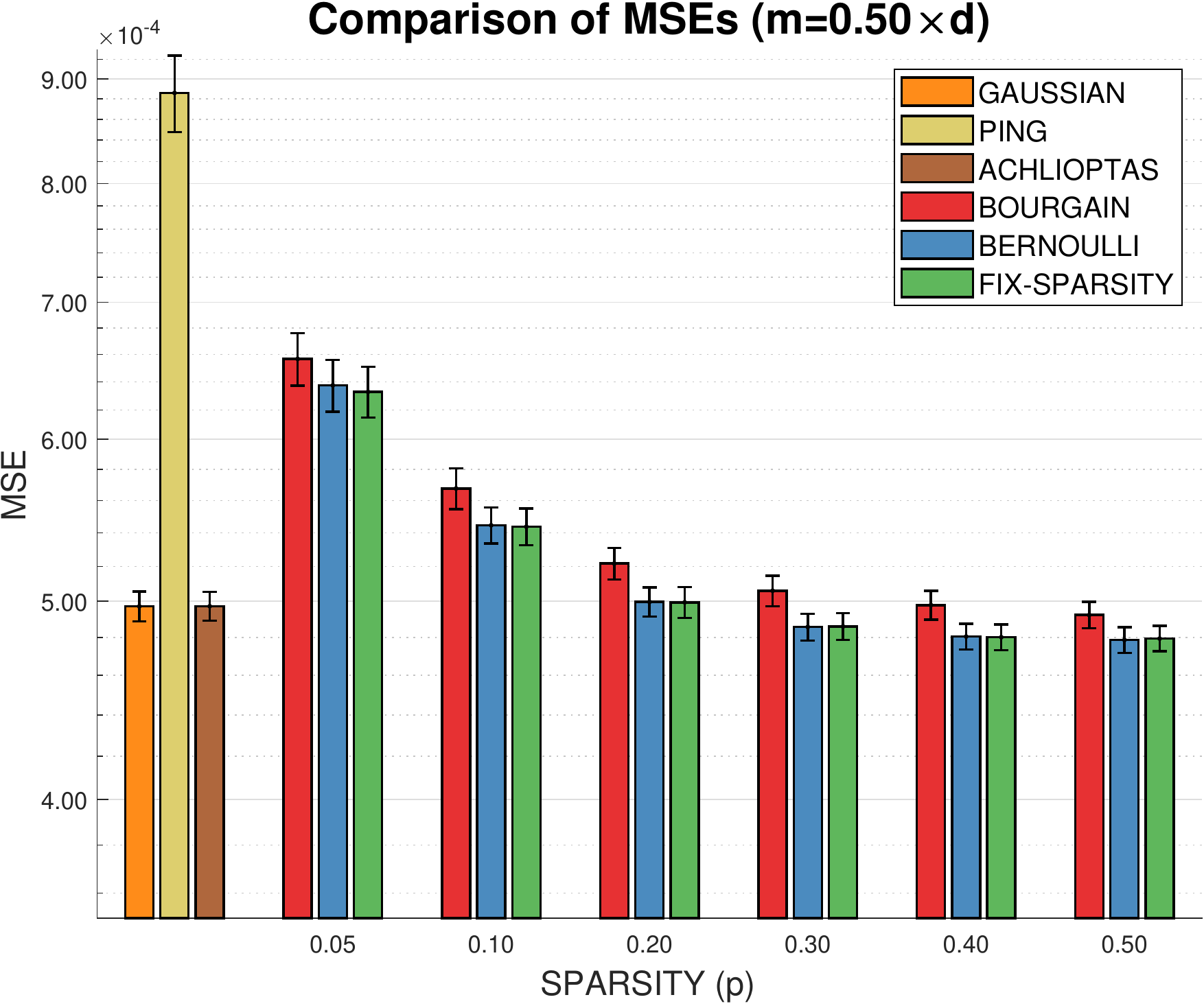}
   \label{fig2:rcv1:m50}
 }

\caption{\textbf{Comparison of Mean Squared Errors}. X-axis: values of $p$; Y-axis: MSEs in log-scale over $1,000$ runs; Error bar: standard deviation of MSEs (a missing bar means the deviation is too large to fit in the figure). The leftmost group in each subfigure gives the results of GAUSSIAN/PING/ACHLIOPTAS that are not dependent on $p$. The rest groups give the results for BOURGAIN/BERNOULLI/FIX-SPARSITY with various $p$ values from $0.05$ to $0.50$.}
\label{fig:mses}
\end{figure*}

\begin{figure*}[!t]
\centering
 \subfigure[GLOVE($m=0.1d$)]{
   \includegraphics[width=1.6in,height=0.9in]{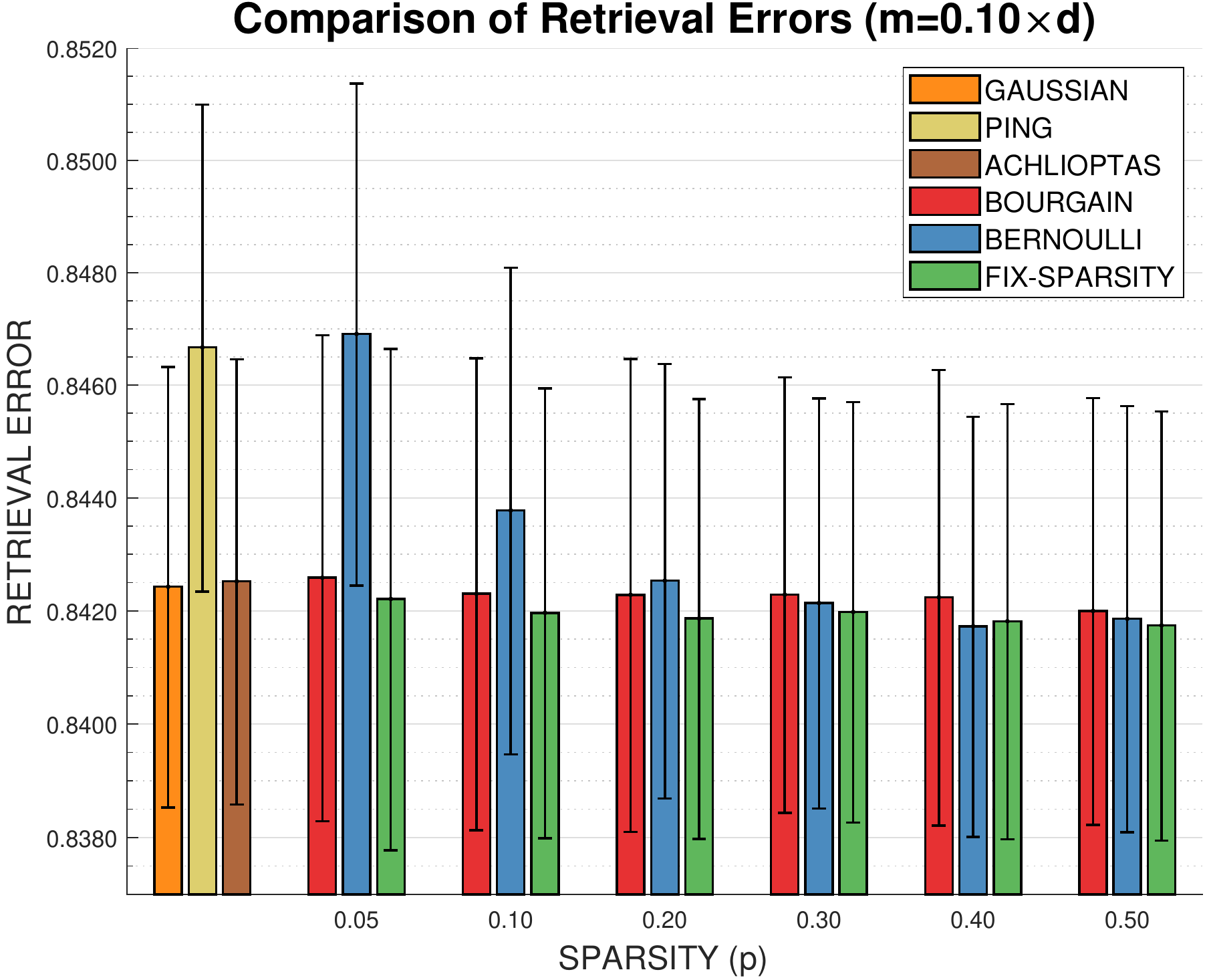}
   \label{fig3:glove:m10}
 }
 \subfigure[IMAGENET($m=0.1d$)]{
   \includegraphics[width=1.6in,height=0.9in]{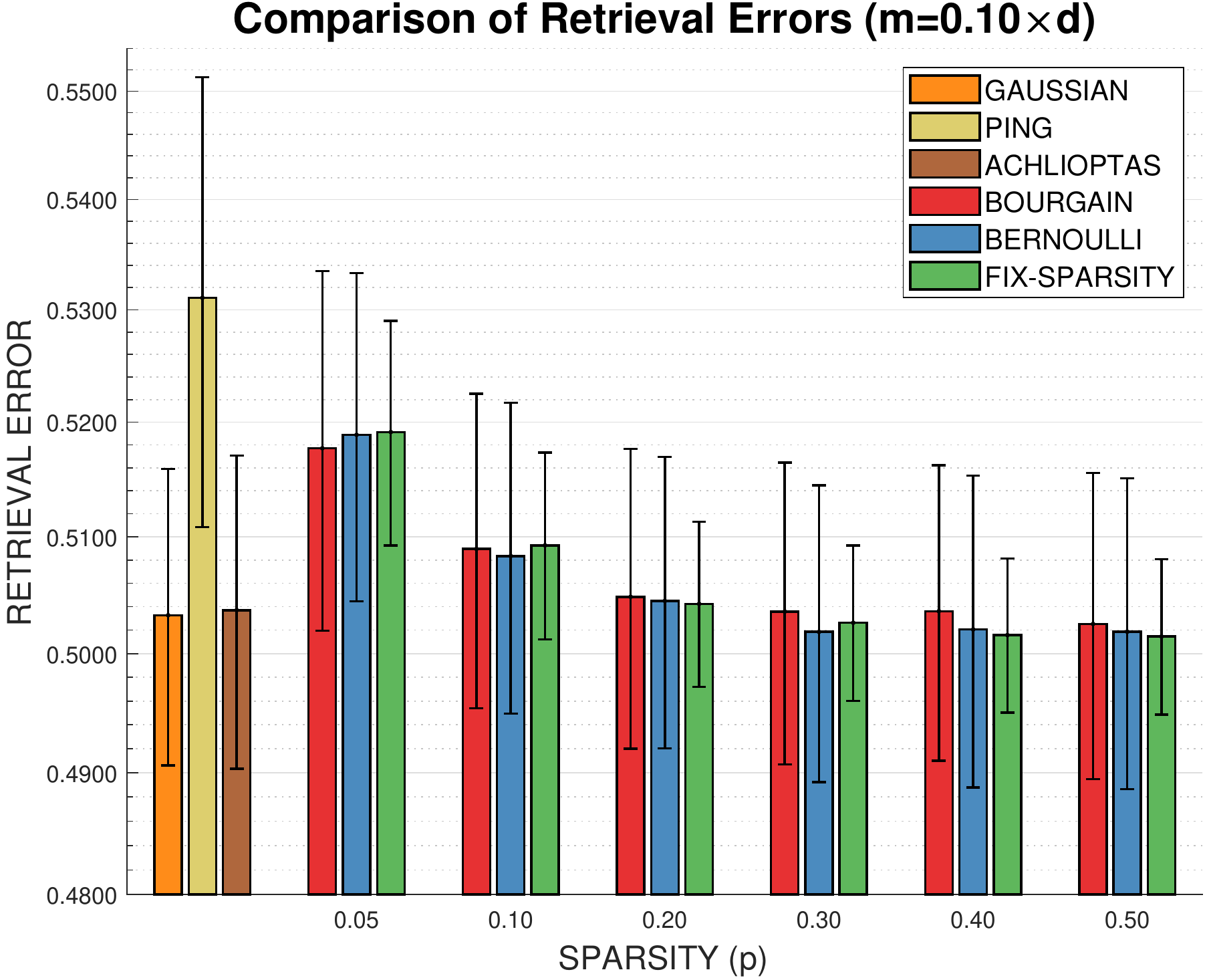}
   \label{fig3:imagenet:m10}
 }
 \subfigure[RCV1($m=0.1d$)]{
   \includegraphics[width=1.6in,height=0.9in]{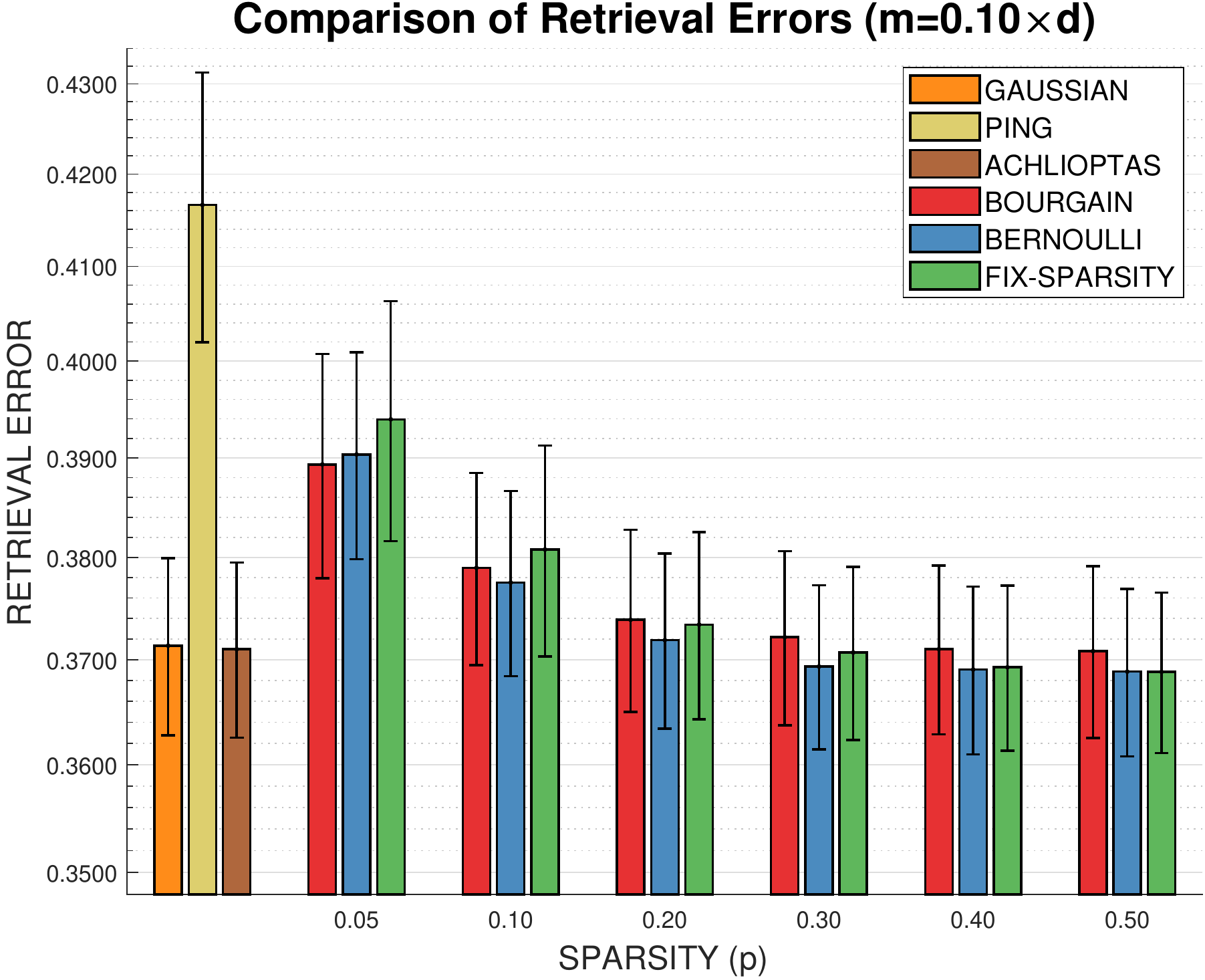}
   \label{fig3:rcv1:m10}
 } \\
 \subfigure[GLOVE($m=0.5d$)]{
   \includegraphics[width=1.6in,height=0.9in]{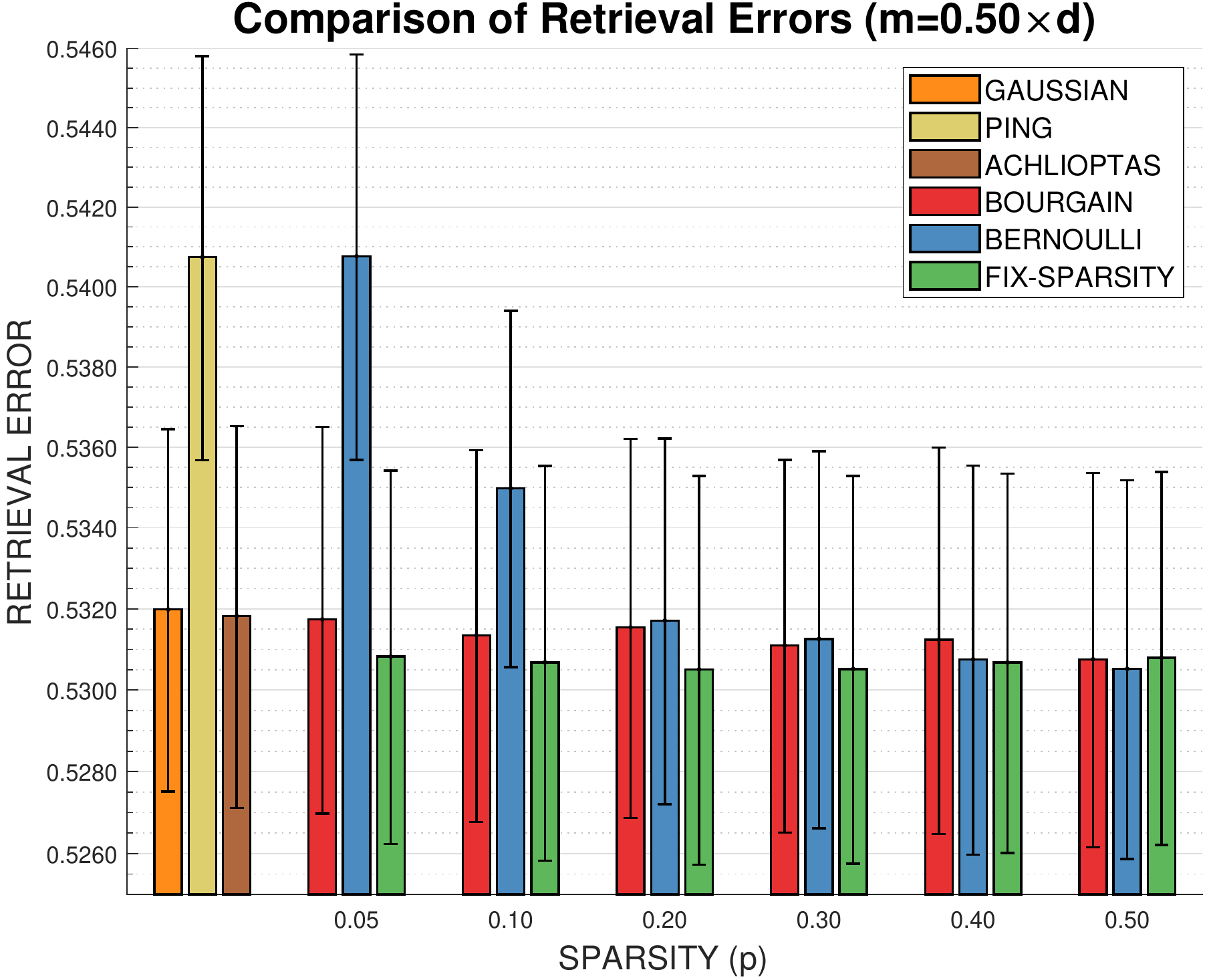}
   \label{fig3:glove:m50}
 }
 \subfigure[IMAGENET($m=0.5d$)]{
   \includegraphics[width=1.6in,height=0.9in]{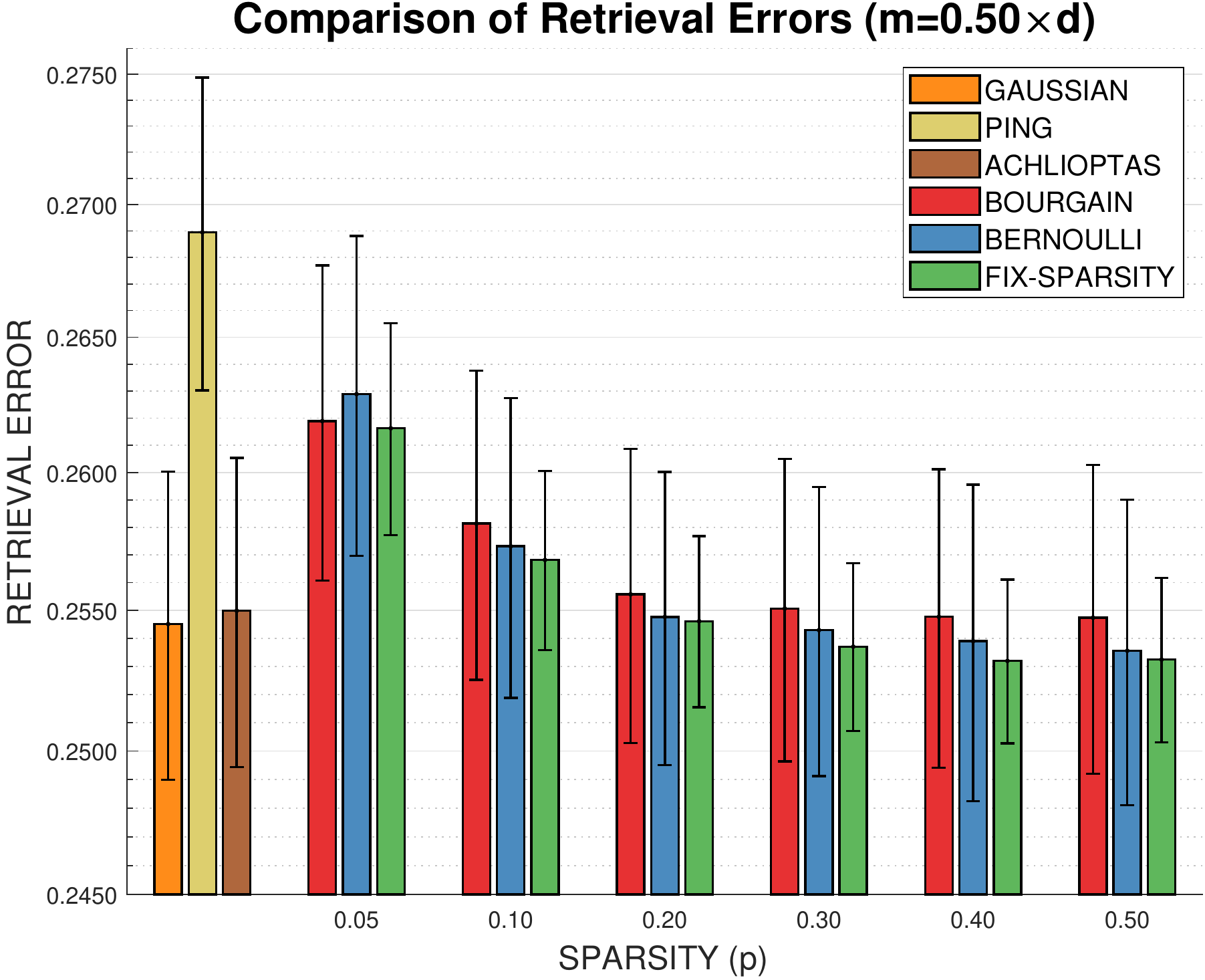}
   \label{fig3:imagenet:m50}
 }
 \subfigure[RCV1($m=0.5d$)]{
   \includegraphics[width=1.6in,height=0.9in]{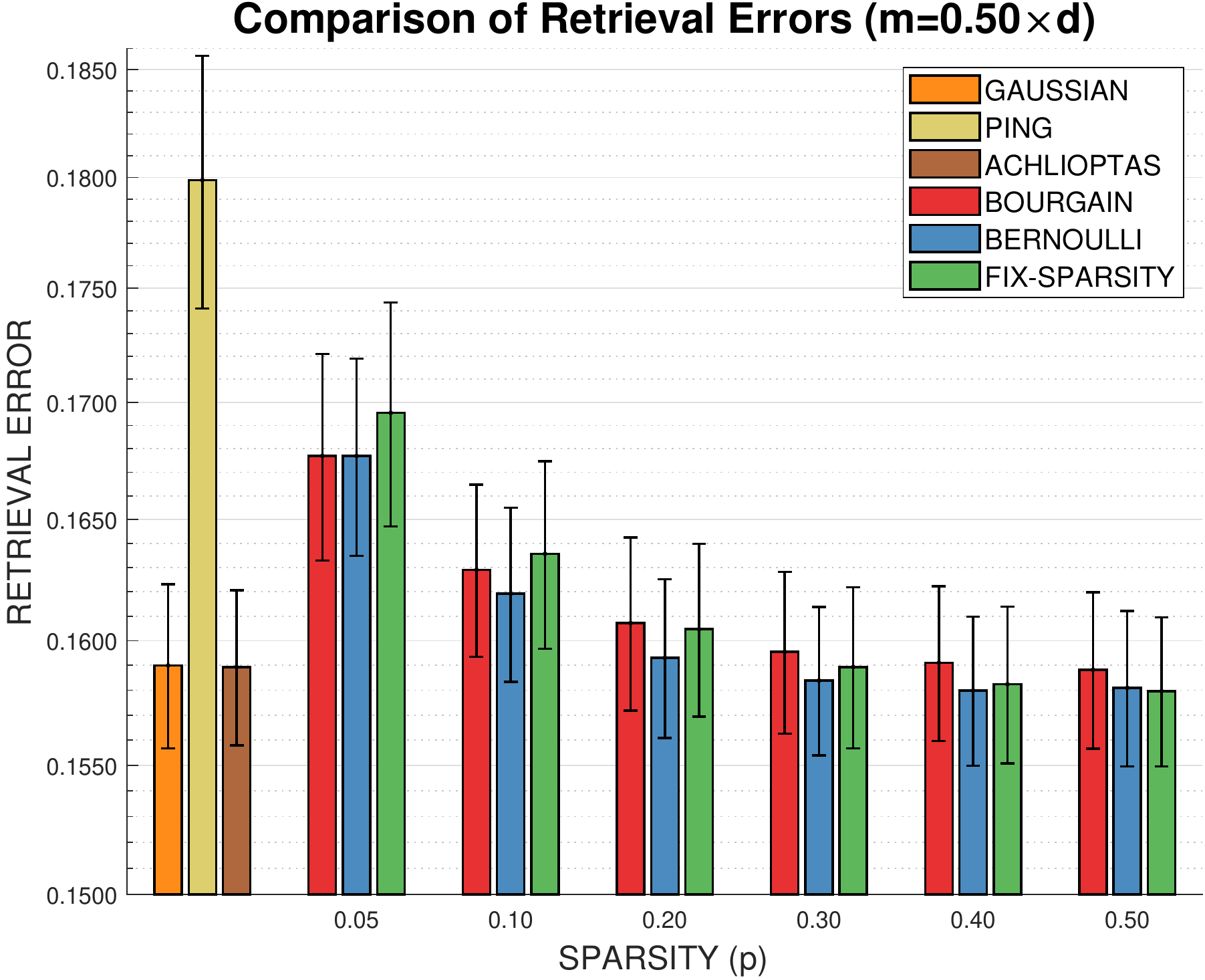}
   \label{fig3:rcv1:m50}
 }

\caption{\textbf{Comparison of Retrieval Errors}. X-axis: values of $p$; Y-axis: mean retrieval errors in log-scale over $1,000$ runs; Error bar: standard deviation of errors. The leftmost group in each subfigure gives the results of GAUSSIAN/PING/ACHLIOPTAS that are not dependent on $p$. The rest groups give the results for  BOURGAIN/BERNOULLI/FIX-SPARSITY with various $p$ values from $0.05$ to $0.50$.}
\label{fig:accs}
\end{figure*}

\subsection{Comparison of variances}
\label{sec:eval:vars}

Based on the analysis in Section \ref{sec:model:moments}, we now compute the variances of the estimates from the GAUSSIAN/BERNOULLI/FIX-SPARSITY approaches. All three methods have closed-form variance formulae.

For each sample, we compute its variance and then report the mean over all samples. The results are shown in Fig. \ref{fig:variances}. It can be seen that, when $p$ increases from $0.05$ to $0.5$, the variance of the estimates for the Bernoulli matrix decreases. When $p\ge 0.3$, the Bernoulli matrix evidently produces smaller variance than that of the Gaussian matrix. The improvement is especially pronounced on the IMAGENET data. Under the setting of $p=0.5$, the variance generated by the Bernoulli matrix is less than one tenth of that generated by the Gaussian matrix. When comparing the variances from the Bernoulli matrix and those from the fixed-sparsity matrix (by setting $c=\lfloor d\times p\rfloor$), typically the fixed-sparsity matrix produces even smaller variances than the Bernoulli matrix does.

\subsection{Comparison of mean squared errors}
\label{sec:eval:mse}

Next we evaluate the distance-preserving capabilities of the proposed models. First, we compute a pairwise distance matrix $D_{d}$ for the $d$-dimensional samples. Then, we project the samples to $m$-dimension by applying different models, with $m=0.1d$ and $0.5d$ respectively. Finally we compute the distance matrix $D_{m}$ in $m$-dimensions. The performance of each approach is measured by the mean squared error (MSE) between $D_{d}$ and $D_{m}$ over $1,000$ runs.

The results are shown in Fig.~\ref{fig:mses}. From the results, we observe that the reported MSEs of the GAUSSIAN/BERNOULLI/FIX-SPARSITY approaches are highly consistent with the variances shown in Fig.~\ref{fig:variances}. The performance of BERNOULLI improves with increasing $p$. When $p\ge 0.3$, the approach consistently reports better MSEs than that of GAUSSIAN.

The BOURGAIN approach, whose projection matrix has exactly $c$ ($=\lfloor d\times p\rfloor$) non-zero elements in each row, also reports improved results when $c$ increases. On the dense GLOVE dataset, BOUGAIN and BERNOULLI have similar performances when $p\ge 0.2$. However, on sparser datasets (IMAGENET and RCV1), the BERNOULLI approach reports better results than BOURGAIN does, statistically significant with a $p$-value of $5\%$ \cite{lehmann2006testing}. Comparatively, the BERNOULLI approach appears to work more easily when projecting sparse vectors.

The performance of FIX-SPARSITY turns out to be less sensitive with the $p$ value as compared to BERNOULLI and BOURGAIN. It reports improved results over BERNOULLI in most settings. On GLOVE dataset, the approach reports excellent performances even with a very small $p=0.05$.

\subsection{Comparison of similarity search errors}
\label{sec:eval:acc}

Having established the improved results in reducing variances and MSEs with the proposed approaches, next we shall further investigate if this improvement does benefit real applications. Specifically, we shall compare the performances of the proposed approaches in similarity search tasks~\cite{manning2008introduction,andoni2008near}.

In each run of the experiment, we map a dataset from $d$-dimension to $m$-dimension with a projection matrix. Then, a sample from the dataset is chosen, in turn, as a testing sample, while the other samples form the candidate set. We obtain the $r$-nearest neighbors ($r=100$) after computing the $m$-dimensional pairwise distances between the testing sample and all candidate samples. These $r$-nearest neighbors are compared against the $r$-nearest neighbors based on the original $d$-dimension and the non-overlapping ratio is taken as the retrieval error of this testing sample. Then, the mean error of all testing samples is used to test the projection approach in one run.

We repeat the experiments for $1,000$ runs and report the results in Fig.~\ref{fig:accs}. Consistent with the results shown in Figs.~\ref{fig:variances} and~\ref{fig:mses}, the Bernoulli matrices report improved results when $p\ge 0.3$, while the fixed-sparsity matrices demonstrate an even slightly better performance. 

%% conclusion
\section{Conclusion}
\label{sec:conc}

Partly inspired by the recent discoveries in neuroscience, in this paper we revisited the classical problem of random projections. We proposed two sparse binary projection models by using Bernoulli random matrices and fixed-sparsity random binary matrices respectively. We established the moments and the concentration bounds for the estimates based on these two projection matrices. Consequently, we proved Johnson--Lindenstrauss type results. Through our theoretical and experimental studies, the benefits of the new models -- both in accuracy and in efficiency -- have been demonstrated. We believe that our investigation helps to shed new lights on the nature of sparse random projection.

%%%%%%%%%%%%%%%%% appendices %%%%%%%%%%%%%%%%%
\newpage
\appendix

\section{The Bernoulli Model}
\label{sec:text:bernoulli}

\subsection{Basic Relationships}
\label{sec:text:bernoulli:basic}

The projection matrix $W_{B}=\left(\xi_{ij}\right)_{m\times d}$ is constructed with each entry $\xi_{ij}$ being generated as i.i.d.\ Bernoulli random variable:
\[
\xi_{ij} = \left\{
\begin{array}{rl}
1, & \mbox{with probability $p$}, \\
0, & \mbox{with probability $\left(1-p\right)$},
\end{array}
\right.
\]
where $0<p \le \frac{1}{2}$, $i=1,...,m$ and $j=1,...,d$.

Consider a given $\boldx=\left( x_{1},\cdots ,x_{d}\right) ^{\top}\in \RR^{d}$. Denote
$\boldeta =\left( \eta _{1},\cdots ,\eta _{m}\right) ^{\top}\in \RR^{m}$ with%
\[
\eta _{i}=\sum_{k=1}^{d}\left( \xi _{ik}-p\right) x_{k}.
\]

Then
\[
\eta _{i}^{2}=\sum_{k=1}^{d}\left( \xi _{ik}-p\right)
^{2}x_{k}^{2}+2\sum_{1\leq k<k^{\prime }\leq d}\left( \xi _{ik}-p\right)
\left( \xi _{ik^{\prime }}-p\right) x_{k}x_{k^{\prime }}.
\]%

We have:
\[
\ex\left( \eta _{i}\right) =0.
\]%
\[
\ex\left( \eta _{i}^{2}\right) =\left[ p\left( 1-p\right) ^{2}+\left(
1-p\right) \left( -p\right) ^{2}\right] \sum_{k=1}^{d}x_{k}^{2}=p\left(
1-p\right) \left\Vert \boldx\right\Vert ^{2}.
\]
\[
\eta _{i}^{2}\leq \left( \sum_{k=1}^{d}\left( 1-p\right) \left\vert
x_{k}\right\vert \right) ^{2}\leq \left( 1-p\right) ^{2}d\left\Vert
\boldx\right\Vert ^{2} .
\]
We also have (cf.~Section \ref{sec:text:bernoulli:fourth}):
\[
\ex\left( \eta _{i}^{4}\right) =p\left( 1-p\right) \left( 1-6p+6p^{2}\right)
\left\Vert \boldx\right\Vert _{4}^{4}+3p^{2}\left( 1-p\right) ^{2}\left\Vert
\boldx\right\Vert ^{4}
\]%
where $\left\Vert \boldx\right\Vert _{4}^{4}=\sum_{k=1}^{d}x_{k}^{4}$ and $%
\left\Vert \boldx\right\Vert ^{4}=\left( \sum_{k=1}^{d}x_{k}^{2}\right) ^{2}$.
Therefore,
\begin{eqnarray*}
\var\left( \left\Vert \boldeta \right\Vert ^{2}\right)
&=&\sum_{i=1}^{m}\var\left( \eta _{i}^{2}\right) =\sum_{i=1}^{m}\left(\ex\left( \eta
_{i}^{4}\right) -\left[ \ex\left( \eta _{i}^{2}\right) \right] ^{2}\right) \\
&=&mp\left( 1-p\right) \left( 1-6p+6p^{2}\right) \left\Vert \boldx\right\Vert
_{4}^{4}+2mp^{2}\left( 1-p\right) ^{2}\left\Vert \boldx\right\Vert ^{4} .
\end{eqnarray*}

\subsection{The Fourth Moment}
\label{sec:text:bernoulli:fourth}

We have
\begin{eqnarray*}
 \var\left( \eta _{i}^{4}\right) &=& \var\left( \sum_{1\leq k\leq d}\left( \xi_{ik}-p\right) x_{k}\right) ^{4} \\
&=& \sum_{1\leq k\leq d}x_{k}^{4}\ex\left( \xi_{ik}-p\right) ^{4}+4\sum_{1\leq
k\neq k^{\prime }\leq d}x_{k}x_{k^{\prime }}^{3}\ex\left[ \left(
\xi_{ik}-p\right) \left( \xi_{ik^{\prime }}-p\right) ^{3}\right]  \\
& & +3\sum_{1\leq k\neq k^{\prime }\leq d}x_{k}^{2}x_{k^{\prime }}^{2}\ex\left[
\left( \xi_{i k}-p\right) ^{2}\left( \xi_{i k^{\prime }}-p\right) ^{2}\right]  \\
& & +6\sum_{1\leq k\neq k^{\prime }\neq \ell \leq d}x_{k}^{2}x_{k^{\prime
}}x_{\ell }\ex\left[ \left( \xi_{i k}-p\right) ^{2}\left( \xi_{i k^{\prime
}}-p\right) \left( \xi_{i \ell }-p\right) \right]  \\
& & +\sum_{1\leq k\neq k^{\prime }\neq \ell \neq \ell ^{\prime }\leq
d}x_{k}x_{k^{\prime }}x_{\ell }x_{\ell ^{\prime }}\ex\left[ \left(
M_{1k}-p\right) \left( \xi_{i k^{\prime }}-p\right) \left( \xi_{i\ell }-p\right)
\left( \xi_{i\ell ^{\prime }}-p\right) \right]  \\
&=& \left[ p\left( 1-p\right) ^{4}
+\left( 1-p\right) p^{4}\right] \sum_{1\leq k\leq d}x_{k}^{4} \\
& & +3\left[ p^{2}\left( 1-p\right) ^{4}+2p\left( 1-p\right) \left( 1-p\right)
^{2}\left( -p\right) ^{2}+\left( 1-p\right) ^{2}\left( -p\right) ^{4}\right]
\sum_{1\leq k\neq k^{\prime }\leq d}x_{k}^{2}x_{k^{\prime }}^{2} \\
&=& p\left( 1-p\right) \left( 1-6p+6p^{2}\right) \left\Vert \boldx\right\Vert
_{4}^{4}+3p^{2}\left( 1-p\right) ^{2}\left\Vert \boldx\right\Vert ^{4}
\end{eqnarray*}
where the third equality is due to the fact that $\xi_{ik}$'s are i.i.d.\ Bernoulli random variables and that $\ex [\xi_{ik}]=p$ for all $i$ and $k$.

\newpage 
\section{The Fixed-Sparsity Model}

\subsection{Basic Relationships}
\label{sec:text:fix:basic}

Consider an experiment of selecting $c$ elements from the $d$ coordinates of the $d$-dimensional vector $\boldx$. There are $\binom{d}{c}$ possible combinations. Denote the index set of each combination by $J_{k}\subset \left\{ 1,\cdots ,d\right\} $ with $\left\vert J_{k}\right\vert =c$, where $k=1,\cdots ,\binom{d}{c}$. Repeat the experiment for $m$ trials independently. Denote by $\xi _{i}=k$ or $I_{\{\xi _{i}=k\}}=1$ if the $i$-th trial chooses the $k$-th combination and $I_{\{\xi _{i}=k\}}=0$ otherwise. Assume the combinations are randomly and uniformly selected, then
\[
\prob \left\{ \xi _{i}=k\right\} =\frac{1}{\binom{d}{c}}
\]%
for all $1\leq i\leq m$ and $1\leq k\leq \binom{d}{c}$. Let the random
binary matrix be generated as
\[
W_{F}:=\left( \xi _{ij}\right) _{m\times d}
\]%
where $\xi _{ij}=1$ if $j\in J_{\xi _{i}}$, and $\xi _{ij}=0$ otherwise.

Consider a given $\boldx=\left( x_{1},\cdots ,x_{d}\right)^\top \in \RR^{d}$. Let $c$ be an integer ($1\leq c\leq \frac{d}{2}$), $q=\frac{1}{d}\left( 1+\sqrt{\frac{d-c}{c\left( d-1\right) }}\right) $, and $s_{x}=\boldone \cdot \boldx$ where $\boldone$ is an all-ones (row) vector.

Define $\boldy\in \RR^{d}$ with $y_{i}=x_{i}-qs_{x}$, for $i=1,...,d$. We have
\begin{equation}
\left\Vert \boldy\right\Vert ^{2}
=\sum_{i=1}^{d}\left( x_{i}-qs_{x}\right) ^{2}
=\left\Vert \boldx\right\Vert ^{2}-2qs_{x}^{2}+dq^{2}s_{x}^{2}
%&=&\left\Vert \boldx\right\Vert ^{2}-\frac{1}{d}\left( 1-\sqrt{\frac{d-c}{c\left(d-1\right) }}\right) \left( 1+\sqrt{\frac{d-c}{c\left( d-1\right) }}\right)s_{x}^{2}  \nonumber \\
=\left\Vert \boldx\right\Vert ^{2}-\frac{c-1}{c\left( d-1\right) }s_{x}^{2}
\leq \left\Vert \boldx\right\Vert ^{2}
\label{x-and-y}
\end{equation}
and
\[
s_{y}
=\boldone \cdot \boldy
=\sum_{j=1}^{d}\left( x_{j}-qs_{x}\right)
=\left(1-dq\right) s_{x}
=-\sqrt{\frac{d-c}{c\left( d-1\right) }}s_{x}.
\]

Denote $\boldeta :=\left( W_{F}-cqE\right) \boldx\in \RR^{m}$, and so
\[
\eta _{i}=\sum_{j\in J_{\xi _{i}}}x_{j}-cqs_{x}=\sum_{j\in J_{\xi
_{i}}}y_{j}.
\]%
Thus, $|\eta _{i}|\leq \sqrt{c}\Vert \boldy\Vert \leq \sqrt{c}\Vert \boldx\Vert $, for all $1\leq i\leq d$. Moreover,
\begin{eqnarray*}
\eta _{i}^{2} &=&\left[ \sum_{1\leq k\leq \binom{d}{c}}I_{\{\xi
_{i}=k\}}\sum_{j\in J_{k}}y_{j}\right] ^{2} \\
&=&\sum_{1\leq k\leq \binom{d}{c}}I_{\{\xi _{i}=k\}}\left( \sum_{j\in
J_{k}}y_{j}\right) ^{2}+\sum_{1\leq k\neq k^{\prime }\leq \binom{d}{c}%
}I_{\{\xi _{i}=k\}}I_{\{\xi _{i}=k^{\prime }\}}\sum_{j\in J_{k},j^{\prime
}\in J_{k^{\prime }}}y_{j}y_{j^{\prime }} \\
&=&\sum_{1\leq k\leq \binom{d}{c}}I_{\{\xi _{i}=k\}}\left( \sum_{j\in
J_{k}}y_{j}\right) ^{2}.
\end{eqnarray*}%
Therefore,
\begin{eqnarray*}
\mathbf{\mathsf{E}}\eta _{i}^{2} &=&\sum_{1\leq k\leq \binom{d}{c}}\prob
\left( \xi _{i}=k\right) \left( \sum_{j\in J_{k}}y_{j}\right) ^{2} \\
&=&\frac{1}{\binom{d}{c}}\left[ \sum_{1\leq j\leq d}\frac{c}{d}\binom{d}{c}%
y_{j}^{2}+\sum_{1\leq j\neq j^{\prime }\leq d}\frac{c\left( c-1\right) }{%
d\left( d-1\right) }\binom{d}{c}y_{j}y_{j^{\prime }}\right]  \\
&=&\frac{c\left( c-1\right) }{d\left( d-1\right) }\left[ \frac{d-1}{c-1}%
\sum_{1\leq j\leq d}y_{j}^{2}+\sum_{1\leq j\neq j^{\prime }\leq
d}y_{j}y_{j^{\prime }}\right]  \\
&=&\frac{c\left( c-1\right) }{d\left( d-1\right) }\left[ \frac{d-c}{c-1}%
\left\Vert \boldy\right\Vert ^{2}+\left( \sum_{1\leq j\leq d}y_{j}\right) ^{2}%
\right]  \\
&=&\frac{c\left( d-c\right) }{d\left( d-1\right) }\left\Vert \boldy\right\Vert
^{2}+\frac{c\left( c-1\right) }{d\left( d-1\right) }s_{y}^{2} \\
&=& \frac{c\left( d-c\right) }{d\left( d-1\right) }\left\Vert \boldx\right\Vert^{2}
\end{eqnarray*}%
where the last equality is due to \eqref{x-and-y}. This shows that $\mathbf{
\mathsf{E}}\eta _{i}^{2}\approx p\left( 1-p\right) \left\Vert \boldx\right\Vert
^{2}$ when $d\gg 1$.

\subsection{The Fourth Moment}
\label{sec:text:fix:fourth}

In this section we shall compute the fourth moment of $\eta _{i}$:
\[
\eta _{i}^{4}=\left( \eta _{i}^{2}\right) ^{2}=\left[ \sum_{1\leq k\leq
\binom{d}{c}}I_{\{\xi _{i}=k\}}\left( \sum_{j\in J_{k}}y_{j}\right) ^{2}%
\right] ^{2}=\sum_{1\leq k\leq \binom{d}{c}}I_{\{\xi _{i}=k\}}\left(
\sum_{j\in J_{k}}y_{j}\right) ^{4}=\left( \sum_{j\in J_{\xi
_{i}}}y_{j}\right) ^{4}
\]%
and so
\[
\mathbf{\mathsf{E}}\eta _{i}^{4}=\sum_{1\leq k\leq \binom{d}{c}} \prob \left(
\xi _{i}=k\right) \left( \sum_{j\in J_{k}}y_{j}\right) ^{4}=\frac{1}{\binom{d%
}{c}}\sum_{1\leq k\leq \binom{d}{c}}\left( \sum_{j\in J_{k}}y_{j}\right)
^{4}.
\]

Let us now compute $\mathbf{\mathsf{E}} \eta _{i}^{4}$. To do this, we need
to expand $\sum_{1\leq k\leq \binom{d}{c}}\left( \sum_{j\in
J_{k}}y_{j}\right) ^{4}$.

In the following, unless otherwise specified, we assume $1\leq k\leq \binom{d%
}{c}$, $j,j^{\prime },\ell ,\ell ^{\prime }\in J_{k}$, and $j<j^{\prime
}<\ell <\ell ^{\prime }$. That is, $\sum_{k}\sum_{j}$ means $\sum_{1\leq
k\leq \binom{d}{c}}\sum_{j\in J_{k}}$, while $\sum_{1\leq j\leq d}$ means
the summation is carried on $j$ from $1$ to $d$. We also assume that $c\ge 5$
and $d\ge 2c$.

We start from:%
\begin{eqnarray}
\sum_{k}\left( \sum_{j}y_{j}\right) ^{4}  \label{equ:expansion_0} 
&=&\sum_{k}\left( \sum_{j}y_{j}^{2}+2\sum_{j<j^{\prime }}y_{j}y_{j^{\prime
}}\right) ^{2}  \nonumber \\
&=&\sum_{k}\left[ \left( \sum_{j}y_{j}^{2}\right) ^{2}+4\left(
\sum_{j}y_{j}^{2}\right) \left( \sum_{j<j^{\prime }}y_{j}y_{j^{\prime
}}\right) +4\left( \sum_{j<j^{\prime }}y_{j}y_{j^{\prime }}\right) ^{2} %
\right] .  \nonumber
\end{eqnarray}

Note that
\begin{equation}
\sum_{k}\left( \sum_{j}y_{j}^{2}\right) ^{2}=\sum_{k}\left[
\sum_{j}y_{j}^{4}+2\sum_{j<j^{\prime }}y_{j}^{2}y_{j^{\prime }}^{2}\right] ,
\label{equ:expansion_1}
\end{equation}
and
\begin{equation}
4\sum_{k}\left[ \left( \sum_{j}y_{j}^{2}\right) \left( \sum_{j<j^{\prime
}}y_{j}y_{j^{\prime }}\right) \right] =4\sum_{k}\left[ \sum_{j<j^{\prime
}}\left( y_{j}^{3}y_{j^{\prime }}+y_{j}y_{j^{\prime }}^{3}\right)
+\sum_{j<j^{\prime }<\ell }\left( y_{j}^{2}y_{j^{\prime }}y_{\ell
}+y_{j}y_{j^{\prime }}^{2}y_{\ell }+y_{j}y_{j^{\prime }}y_{\ell }^{2}\right) %
\right]  \label{equ:expansion_2}
\end{equation}
and
\begin{equation}
4\sum_{k}\left( \sum_{j<j^{\prime }}y_{j}y_{j^{\prime }}\right)
^{2}=4\sum_{k}\left[ \sum_{j<j^{\prime }}y_{j}^{2}y_{j^{\prime
}}^{2}+6\sum_{j<j^{\prime }<\ell <\ell ^{\prime }}y_{j}y_{j^{\prime
}}y_{\ell }y_{\ell ^{\prime }}+2\sum_{j<j^{\prime }<\ell }\left(
y_{j}^{2}y_{j^{\prime }}y_{\ell }+y_{j}y_{j^{\prime }}^{2}y_{\ell
}+y_{j}y_{j^{\prime }}y_{\ell }^{2}\right) \right] . \label{equ:expansion_3}
\end{equation}

Adding (\ref{equ:expansion_1}), (\ref{equ:expansion_2}) and (\ref{equ:expansion_3}) up yields
\begin{equation}
\sum_{k}\left[
\begin{array}{c}
\sum_{j}y_{j}^{4}+6\sum_{j<j^{\prime }}y_{j}^{2}y_{j^{\prime
}}^{2}+4\sum_{j<j^{\prime }}\left( y_{j}^{3}y_{j^{\prime
}}+y_{j}y_{j^{\prime }}^{3}\right) \\
+12\sum_{j<j^{\prime }<\ell }\left( y_{j}^{2}y_{j^{\prime }}y_{\ell
}+y_{j}y_{j^{\prime }}^{2}y_{\ell }+y_{j}y_{j^{\prime }}y_{\ell }^{2}\right)
+24\sum_{j<j^{\prime }<\ell <\ell ^{\prime }}y_{j}y_{j^{\prime }}y_{\ell
}y_{\ell ^{\prime }}
\end{array}
\right]  . \label{equ:expansion_4}
\end{equation}

Let us examine the terms in (\ref{equ:expansion_4}). First,
\begin{equation}
\sum_{k}\sum_{j}y_{j}^{4}=\frac{c}{d}\binom{d}{c}\left\Vert \boldy\right\Vert
_{4}^{4}=\binom{d-1}{c-1}\left\Vert \boldy\right\Vert _{4}^{4},
\label{equ:expansion_5}
\end{equation}
and
\begin{eqnarray}
6\sum_{k}\sum_{j<j^{\prime }}y_{j}^{2}y_{j^{\prime }}^{2} &=&6\binom{d-2}{c-2%
}\sum_{1\leq j<j^{\prime }\leq d}y_{j}^{2}y_{j^{\prime }}^{2}  \nonumber \\
&=&6\binom{d-2}{c-2}\frac{\left( \sum_{1\leq j\leq d}y_{j}^{2}\right)
^{2}-\sum_{1\leq j\leq d}y_{j}^{4}}{2}  \nonumber \\
&=&3\binom{d-2}{c-2}\left( \left\Vert \boldy\right\Vert ^{4}-\left\Vert
\boldy\right\Vert _{4}^{4}\right) ,  \label{equ:expansion_6}
\end{eqnarray}
and
\begin{eqnarray}
4\sum_{k}\sum_{j<j^{\prime }}\left( y_{j}^{3}y_{j^{\prime
}}+y_{j}y_{j^{\prime }}^{3}\right) &=&4\binom{d-2}{c-2}\sum_{1\leq j\neq
j^{\prime }\leq d}y_{j}y_{j^{\prime }}^{3}  \nonumber \\
&=&4\binom{d-2}{c-2}\left( \sum_{1\leq j\leq d}y_{j}^{3}\sum_{1\leq
j^{\prime }\leq d}^{d}y_{j^{\prime }}-\sum_{1\leq j\leq
d}^{d}y_{j}^{4}\right)  \nonumber \\
&=&4\binom{d-2}{c-2}\left( \left\Vert \boldy\right\Vert _{3}^{3}s_{y}-\left\Vert \boldy\right\Vert _{4}^{4}\right),
\label{equ:expansion_7}
\end{eqnarray}
and
\begin{eqnarray}
&&12\sum_{k}\sum_{j<j^{\prime }<\ell }\left( y_{j}^{2}y_{j^{\prime }}y_{\ell
}+y_{j}y_{j^{\prime }}^{2}y_{\ell }+y_{j}y_{j^{\prime }}y_{\ell }^{2}\right)
\nonumber \\
&=&12\binom{d-3}{c-3}\left[ \sum_{1\leq j\leq d}y_{j}^{2}\sum_{1\leq
j^{\prime }<\ell \leq d}y_{j^{\prime }}y_{\ell }-\sum_{1\leq j\neq j^{\prime
}\leq d}y_{j}y_{j^{\prime }}^{3}\right]  \nonumber \\
&=&12\binom{d-3}{c-3}\left[ \left\Vert \boldy\right\Vert ^{2}\frac{\left(
\sum_{1\leq j\leq d}y_{j}\right) ^{2}-\sum_{1\leq j\leq d}y_{j}^{2}}{2}%
-\left( \left\Vert \boldy\right\Vert _{3}^{3}s_{y}-\left\Vert \boldy\right\Vert
_{4}^{4}\right) \right]  \nonumber \\
&=&12\binom{d-3}{c-3}\left[ \frac{1}{2}\left\Vert \boldy\right\Vert ^{2}s_{y}^{2}-%
\frac{1}{2}\left\Vert \boldy\right\Vert ^{4}-\left\Vert \boldy\right\Vert
_{3}^{3}s_{y}+\left\Vert \boldy\right\Vert _{4}^{4}\right]  \nonumber \\
&=&\binom{d-3}{c-3}\left[ 6\left\Vert \boldy\right\Vert ^{2}s_{y}^{2}-6\left\Vert
\boldy\right\Vert ^{4}-12\left\Vert \boldy\right\Vert _{3}^{3}s_{y}+12\left\Vert
\boldy\right\Vert _{4}^{4}\right] ,  \label{equ:expansion_8}
\end{eqnarray}
and
\begin{eqnarray}
&&24\sum_{k}\sum_{j<j^{\prime }<\ell <\ell ^{\prime }}y_{j}y_{j^{\prime
}}y_{\ell }y_{\ell ^{\prime }}  \nonumber \\
&=&24\binom{d-4}{c-4}\sum_{1\leq j<j^{\prime }<\ell <\ell ^{\prime }\leq
d}y_{j}y_{j^{\prime }}y_{\ell }y_{\ell ^{\prime }}  \nonumber \\
&=&4\binom{d-4}{c-4}\left[ \left( \sum_{1\leq j<j^{\prime }\leq
d}y_{j}y_{j^{\prime }}\right) ^{2}-\sum_{1\leq j<j^{\prime }\leq
d}y_{j}^{2}y_{j^{\prime }}^{2}-2\sum_{1\leq j<j^{\prime }<\ell \leq d}\left(
y_{j}^{2}y_{j^{\prime }}y_{\ell }+y_{j}y_{j^{\prime }}^{2}y_{\ell
}+y_{j}y_{j^{\prime }}y_{\ell }^{2}\right) \right]  \nonumber \\
&=&4\binom{d-4}{c-4}\left[ \left( \frac{s_{y}^{2}-\left\Vert \boldy\right\Vert
^{2}}{2}\right) ^{2}-\frac{\left\Vert \boldy\right\Vert ^{4}-\left\Vert
\boldy\right\Vert _{4}^{4}}{2}-2\left( \frac{1}{2}\left\Vert \boldy\right\Vert
^{2}s_{y}^{2}-\frac{1}{2}\left\Vert \boldy\right\Vert ^{4}-\left\Vert
\boldy\right\Vert _{3}^{3}s_{y}+\left\Vert \boldy\right\Vert _{4}^{4}\right) \right]
\nonumber \\
&=&4\binom{d-4}{c-4}\left[ \frac{s_{y}^{4}+\left\Vert \boldy\right\Vert
^{4}-2\left\Vert \boldy\right\Vert ^{2}s_{y}^{2}}{4}-\frac{\left\Vert
\boldy\right\Vert ^{4}-\left\Vert \boldy\right\Vert _{4}^{4}}{2}-\left\Vert
\boldy\right\Vert ^{2}s_{y}^{2}+\left\Vert \boldy\right\Vert ^{4}+2\left\Vert
\boldy\right\Vert _{3}^{3}s_{y}-2\left\Vert \boldy\right\Vert _{4}^{4}\right]
\nonumber \\
&=&4\binom{d-4}{c-4}\left[ \frac{3}{4}\left\Vert \boldy\right\Vert ^{4}-\frac{3}{2}\left\Vert \boldy\right\Vert _{4}^{4}+\frac{1}{4}s_{y}^{4}+2\left\Vert
\boldy\right\Vert _{3}^{3}s_{y}-\frac{3}{2}\left\Vert \boldy\right\Vert ^{2}s_{y}^{2}\right]  \nonumber \\
&=&\binom{d-4}{c-4}\left[ 3\left\Vert \boldy\right\Vert ^{4}-6\left\Vert
\boldy\right\Vert _{4}^{4}+s_{y}^{4}+8\left\Vert \boldy\right\Vert
_{3}^{3}s_{y}-6\left\Vert \boldy\right\Vert ^{2}s_{y}^{2}\right] .
\label{equ:expansion_9}
\end{eqnarray}

Summing up equations from (\ref{equ:expansion_5}) to (\ref{equ:expansion_9}%
), we arrive at the following findings:

\begin{itemize}
\item The coefficients for $\left\Vert \boldy\right\Vert ^{4}$ is $3\binom{d-2}{c-2}-6\binom{d-3}{c-3}+3\binom{d-4}{c-4}$.
Note that $\left\Vert \boldy\right\Vert ^{4}\leq \left\Vert \boldx\right\Vert ^{4}$. We have
\begin{eqnarray}
&&\left[ 3\binom{d-2}{c-2}-6\binom{d-3}{c-3}+3\binom{d-4}{c-4}\right]
\left\Vert \boldy\right\Vert ^{4}  \nonumber \\
&\leq &\left[ 3\binom{d-2}{c-2}-6\binom{d-3}{c-3}+3\binom{d-4}{c-4}\right]
\left\Vert \boldx\right\Vert ^{4}  \label{equ:expansion_10}
\end{eqnarray}

\item The coefficients for $\left\Vert \boldy\right\Vert _{4}^{4}$ is $\binom{d-1}{c-1}-7\binom{d-2}{c-2}+12\binom{d-3}{c-3}-6\binom{d-4}{c-4}$. Note that
$0\leq \left\Vert \boldy\right\Vert _{4}^{4}\leq \left\Vert \boldy\right\Vert ^{4}\leq \left\Vert \boldx\right\Vert ^{4}$ and $5\le c\le \frac{d}{2}$. We have%
\begin{eqnarray}
&&\left[ \binom{d-1}{c-1}-7\binom{d-2}{c-2}+12\binom{d-3}{c-3}-6\binom{d-4}{%
c-4}\right] \left\Vert \boldy\right\Vert _{4}^{4}  \nonumber \\
&<&\left[ \binom{d-1}{c-1}-2\binom{d-2}{c-2}\right] \left\Vert \boldy\right\Vert_{4}^{4}  \nonumber \\
&\leq &\left[ \binom{d-1}{c-1}-2\binom{d-2}{c-2}\right] \left\Vert
\boldy\right\Vert ^{4}  \nonumber \\
&\leq &\left[ \binom{d-1}{c-1}-2\binom{d-2}{c-2}\right] \left\Vert
\boldx\right\Vert ^{4} .  \label{equ:expansion_11}
\end{eqnarray}

\item The coefficients for $s_{y}^{4}$ is $\binom{d-4}{c-4}$. Note that $%
s_{y}^{4}\leq \frac{d^{2}\left( d-c\right) ^{2}}{c^{2}\left( d-1\right) ^{2}}%
\left\Vert \boldx\right\Vert ^{4}$. We have%
\begin{equation}
\binom{d-4}{c-4}s_{y}^{4}\leq \binom{d-4}{c-4}\frac{d^{2}\left( d-c\right)
^{2}}{c^{2}\left( d-1\right) ^{2}}\left\Vert \boldx\right\Vert ^{4}\leq \binom{d-4}{c-4}\frac{\left( d-2\right) \left( d-3\right) }{\left( c-2\right) \left(
c-3\right) }\left\Vert \boldx\right\Vert ^{4}\leq \binom{d-2}{c-2}\left\Vert
\boldx\right\Vert ^{4} .  \label{equ:expansion_12}
\end{equation}

\item The coefficients for $\left\Vert \boldy\right\Vert _{3}^{3}s_{y}$ is $4%
\binom{d-2}{c-2}-12\binom{d-3}{c-3}+8\binom{d-4}{c-4}$. Note that the
coefficient is positive when $d\geq 2c$ and
\[
\left\Vert \boldy\right\Vert _{3}^{3}s_{y}\leq \left( \left\Vert \boldy\right\Vert
^{2}\right) ^{\frac{3}{2}}\left\vert s_{y}\right\vert \leq \left\Vert
\boldy\right\Vert ^{3}\sqrt{\frac{d-c}{c\left( d-1\right) }}\left\vert
s_{x}\right\vert \leq \left\Vert \boldx\right\Vert ^{3}\sqrt{\frac{d-c}{c\left(
d-1\right) }}\sqrt{d}\left\Vert \boldx\right\Vert \leq \sqrt{\frac{d}{c}}%
\left\Vert \boldx\right\Vert ^{4}.
\]%
We have
\begin{eqnarray}
&&\left[ 4\binom{d-2}{c-2}-12\binom{d-3}{c-3}+8\binom{d-4}{c-4}\right]
\left\Vert \boldy\right\Vert _{3}^{3}s_{y}  \nonumber \\
&\leq &\left[ 4\binom{d-2}{c-2}-12\binom{d-3}{c-3}+8\binom{d-4}{c-4}\right]
\sqrt{\frac{d}{c}}\left\Vert \boldx\right\Vert ^{4}  \nonumber \\
&\leq &\left[ 4\times \frac{d-1}{c-1}\binom{d-2}{c-2}-12\times \frac{2}{3}%
\times \frac{d-2}{c-2}\binom{d-3}{c-3}+8\times \frac{3}{4}\times \frac{d-3}{%
c-3}\binom{d-4}{c-4}\right] \left\Vert \boldx\right\Vert ^{4}  \nonumber \\
&\leq &\left[ 4\binom{d-1}{c-1}-8\binom{d-2}{c-2}+6\binom{d-3}{c-3}\right]
\left\Vert \boldx\right\Vert ^{4}.  \label{equ:expansion_13}
\end{eqnarray}

\item The coefficients for $\left\Vert \boldy\right\Vert ^{2}s_{y}^{2}$ is $6%
\binom{d-3}{c-3}-6\binom{d-4}{c-4}$. Note that $\left\Vert
\boldy\right\Vert^{2}s_{y}^{2} \leq \frac{d\left( d-c\right) }{c\left( d-1\right)
}\left\Vert \boldx\right\Vert ^{4} \le \frac{d-2}{c-2}\left\Vert \boldx\right\Vert
^{4} $ and $5\le c\le \frac{d}{2}$. We have
\begin{eqnarray}
&& \left[ 6\binom{d-3}{c-3}-6\binom{d-4}{c-4}\right] \left\Vert
\boldy\right\Vert^{2}s_{y}^{2}  \nonumber \\
&\leq & \left[6\times \frac{d-2}{c-2} \binom{d-3}{c-3}-6\times \frac{1}{2}%
\times \frac{d-3}{c-3}\binom{d-4}{c-4}\right]\left\Vert \boldx\right\Vert ^{4}
\nonumber \\
&= & \left[6\binom{d-2}{c-2} -3\binom{d-3}{c-3}\right] \left\Vert
\boldx\right\Vert ^{4} .  \label{equ:expansion_14}
\end{eqnarray}
\end{itemize}

Summing up equations (\ref{equ:expansion_10}) to (\ref{equ:expansion_14}),
we have
\begin{eqnarray}
\sum_{k}\left( \sum_{j}y_{j}\right) ^{4} &\leq &\left[ 5\binom{d-1}{c-1}-3%
\binom{d-3}{c-3}+3\binom{d-4}{c-4}\right] \left\Vert \boldx\right\Vert ^{4}
\nonumber \\
&\leq &5\binom{d-1}{c-1}\left\Vert \boldx\right\Vert ^{4} ,
\label{equ:summation_bound}
\end{eqnarray}
leading to
\begin{equation}
\mathbf{\mathsf{E}} \eta _{i}^{4}=\frac{1}{\binom{d}{c}}\sum_{k}\left(
\sum_{j\in J_{k}}y_{j}\right) ^{4}\leq \frac{5c}{d}\left\Vert \boldx\right\Vert
^{4} .  \label{equ:expecation_bound}
\end{equation}

\newpage
\section{Proof of Lemma~2}
\label{sec:text:lemma2}

Denote $\sigma _{i}^{2}=\mathbf{\mathsf{Var}}(Z_{i})$, and for any given $s>0
$:
\begin{eqnarray}
\prob \left\{ S\geq t \right\}
&=&\prob \left\{ \sum_{i=1}^{m}s(Z_{i}-\ex Z_{i})\geq s t \right\}   \nonumber \\
&=&\prob \left\{ \exp \left( \sum_{i=1}^{m}s(Z_{i}-\ex Z_{i})\right) \geq \exp \left( s t \right) \right\}   \nonumber \\
&\leq &\frac{\ex\left( \exp \left( \sum_{i=1}^{m}s(Z_{i}-\ex Z_{i})\right) \right) }{\exp (s t )}  \nonumber \\
&=&\frac{\prod_{i=1}^{m}\ex (\exp (s(Z_{i}-\ex Z_{i})))}{\exp (s t)}
\label{equ:lemma2:dev-prob}
\end{eqnarray}%
where the third step is due to Markov's inequality. Note that $|Z_{i}-\ex Z_{i}|\leq b$, and
\begin{eqnarray*}
\ex (\exp (s(Z_{i}-\ex Z_{i})))
&=&\ex \left( \sum_{k=0}^{\infty }\frac{s^{k}(Z_{i}-\ex Z_{i})^{k}}{k!}\right)  \\
&\leq &1+\sum_{k=2}^{\infty }\frac{s^{k}\ex (Z_{i}-\ex Z_{i})^{2}b^{k-2}}{k!} \\
&=&1+\frac{\sigma _{i}^{2}}{b^{2}}\sum_{i=2}^{\infty }\frac{s^{k}b^{k}}{k!}
\\
&=&1+\frac{\sigma _{i}^{2}}{b^{2}}\left( e^{sb}-1-sb\right)  \\
&\leq &\exp \left( \frac{\sigma _{i}^{2}}{b^{2}}\left( e^{sb}-1-sb\right)
\right) .
\end{eqnarray*}%
Applying the above inequality to \eqref{equ:lemma2:dev-prob} yields
\begin{eqnarray}
\prob \left\{ S\geq t \right\}  &\leq &\exp \left( \frac{\sum_{i=1}^{m}\sigma_{i}^{2}}{b^{2}}\left( e^{sb}-1-sb\right) -s t\right)   \nonumber \\
&=&\exp \left( \frac{\sum_{i=1}^{m}(\mathbf{\mathsf{E}}X_{i}^{2}-(\mathbf{\mathsf{E}}X_{i})^{2})}{b^{2}}\left( e^{sb}-1-sb\right) -s t\right)
\nonumber \\
&\leq &\exp \left( \frac{w}{b^{2}}\left( e^{sb}-1-sb\right) -s t\right) .
\label{equ:lemma2:bennett-ineq}
\end{eqnarray}%
Since the minimizer of $\frac{w}{b^{2}}\left( e^{sb}-1-sb\right) -s t$ over $%
s>0$ is $s^{\ast }=\frac{1}{b}\ln \left( 1+\frac{bt}{w}\right) $,
substituting $s=s^{\ast }$ into \eqref{equ:lemma2:bennett-ineq} yields the desired
inequality.

\newpage
\section{Proof of Lemma~5}
\label{sec:text:lemma5}

Let $Z_{i}=\eta _{i}^{2}$, $b=c\left\Vert \boldx\right\Vert ^{2}$,
$w=5m\frac{c}{d}\left\Vert \boldx\right\Vert ^{4}$ and
$t=\epsilon m\frac{c\left( d-c\right) }{d\left( d-1\right) }\left\Vert \boldx\right\Vert ^{2}$.
When $\epsilon <\frac{20}{c}$, we have $\frac{bt}{w}<4$ and $0.25\left( \frac{bt}{w}\right)
^{2}<h\left( \frac{bt}{w}\right) $. Applying Bennett's inequality, we have
\begin{eqnarray*}
&&\prob \left\{ \sum_{i=1}^{m}\left( \eta _{i}^{2}-\mathbf{\mathsf{E}}\eta
_{i}^{2}\right) \geq \epsilon \sum_{i=1}^{m}\mathbf{\mathsf{E}}\eta
_{i}^{2}\right\}  \\
&\leq &\exp \left( -\frac{t^{2}}{4w}\right)  \\
&=&\exp \left( -\frac{\epsilon ^{2}m^{2}\left( \frac{c\left( d-c\right) }{%
d\left( d-1\right) }\right) ^{2}\left\Vert \boldx\right\Vert ^{4}}{20m\frac{c}{d}%
\left\Vert \boldx\right\Vert ^{4}}\right)  \\
&<&\exp \left( -\frac{\epsilon ^{2}m\frac{c}{d}\left( \frac{d-c}{d}\right)
^{2}}{20}\right)  \\
&=&\exp \left( -\frac{\epsilon ^{2}mp\left( 1-p\right) ^{2}}{20}\right)
\end{eqnarray*}%
where $0<p=\frac{c}{d}\leq \frac{1}{2}$. Lemma~5 is thus proven.

\end{document}